\newif\ifdraft
\draftfalse

\documentclass{article} 
\usepackage{iclr2027_conference,times}
\iclrpreprintcopy

\usepackage{amsmath,amsfonts,bm}

\def\eqref#1{equation~\ref{#1}}

\def\1{\bm{1}}

\DeclareMathAlphabet{\mathsfit}{\encodingdefault}{\sfdefault}{m}{sl}
\SetMathAlphabet{\mathsfit}{bold}{\encodingdefault}{\sfdefault}{bx}{n}

\usepackage{amsmath,amssymb}
\usepackage{hyperref}
\usepackage{microtype}
\usepackage{url}
\usepackage{float}
\usepackage{graphicx}
\usepackage{booktabs}
\usepackage{array}
\usepackage[section]{placeins}
\usepackage{flafter}
\usepackage{comment}
\usepackage{subcaption}

\usepackage{tabularx}
\usepackage{makecell}
\usepackage{threeparttable}
\usepackage{multirow}

\title{Beyond Linear Concepts: Discovering and Aligning Non-Linear Concept Manifolds in Large Language Models}

\author{
Tido Specht$^{1}$ \quad
Elias Benedict Krey$^{2}$ \quad
Nils Neukirch$^{2}$ \quad
Nils Strodthoff$^{2}$ \\[0.25em]
{\small
$^{1}$University of Groningen \qquad
$^{2}$Carl von Ossietzky University of Oldenburg}
}

\usepackage[normalem]{ulem}
\newcommand{\stkout}[1]{\ifmmode\text{\sout{\ensuremath{#1}}}\else\sout{#1}\fi}

\ifdraft

\newcommand{\deleted}[1]{\textcolor{red}{\stkout{#1}}}

\newcommand{\deletedfloat}[1]{}
\newcommand{\commented}[1]{\textcolor{blue}{#1}}
\else

\newcommand{\deleted}[1]{}

\newcommand{\deletedfloat}[1]{}
\newcommand{\commented}[1]{}
\fi

\begin{document}

\maketitle

\begin{abstract}
Understanding information processing in large language models (LLMs) requires dissecting the geometric organization of their internal token representations. While existing mechanistic interpretability (MI) methods seek to extract human-understandable concepts, they are constrained by a strong linearity assumption—an assumption increasingly challenged by evidence of non-linear feature manifolds. In this work, we move beyond linear concepts by adapting Non-Linear Multi-Dimensional Concept Discovery (NLMCD) from computer vision to token-level LLM activations, modeling concepts as low-dimensional manifolds. To enable quantitative comparison of these independently discovered concept manifolds across layers and models, we introduce a concept-based alignment (CBA) score, a generalized Rand index that measures geometric proximity without requiring explicit feature matching. Our analysis yields six key findings: (i) a neighboring-layer sanity check shows CBA is more sensitive than PCA- or CKA-based linear baselines; (ii) layer×layer alignment matrices reveal two characteristic block structures concentrated in intermediate and late layers, consistent across models and obscured by linear metrics; (iii) concept composition remains syntax-dominated through most of the network before giving way to increasingly mixed syntactic–semantic concepts in later layers, with increasing output-orientation toward the final layers; (iv) multilingual concept sharing between English and Mandarin is training-dependent rather than universal — strongest in Qwen, weaker in Llama, and absent in GPT-2; (v) inter-model alignment mirrors this structure: strong correspondence between same-family Qwen models of different scale, consistent with the block structures in (ii), but weak alignment across model families; and (vi) tracing representational alignment across the training stages of the Tulu-3 pipeline shows alignment is highest between adjacent stages, validating CBA, with the largest shift occurring between the base model and the SFT stage, while subsequent preference-alignment stages (DPO, RLVR) leave early layers largely unchanged and RLVR mostly preserves DPO's concepts in late layers.
\end{abstract}

\section{Introduction}
Large Language Models (LLMs) are gaining traction in a variety of tasks, most notably as assistant chatbots, as agents that solve complex tasks, and as coding support, making them among the most influential machine-learning paradigms of the present day. As LLMs are trained, scaled, finetuned, and adapted across languages, a recurring question is whether and how their internal representations change: \emph{Does finetuning reorganize concepts or merely refine them? Do larger models within a family preserve the representational structure of smaller ones? Do multilingual models encode shared, language-agnostic concepts, or largely separate ones per language?} Answering such questions requires comparing internal representations across layers, model scales, training stages, and languages — a task that depends entirely on having a reliable notion of representational alignment.

The standard tools for this comparison are global alignment metrics such as Centered Kernel Alignment (CKA) or canonical correlation analysis (CCA). These methods compare entire activation spaces under an implicit linear representation hypothesis: similarity is measured via linear projections or kernel similarities over the full activation vector, rather than over the discrete, interpretable units (concepts, features) that individual activations may encode. This is a poor fit for evidence that some computationally relevant structures in LLMs are intrinsically multi-dimensional and non-linear \citep{engels2024nonlinear}, and it makes such metrics comparatively insensitive: two representation spaces can appear highly similar under CKA/CCA even when their underlying concept structure diverges substantially at a finer grain.

We investigate an alternative based on \emph{Non-Linear Multi-Dimensional Concept Discovery (NLMCD)} \citep{vielhaben2026beyond}, which treats concepts as density-supported manifolds rather than linear directions. It builds on evidence, including from sparse autoencoders (SAEs), that concept-level decompositions of LLM activations are a productive unit of comparison \citep{anthropic,gao2024scaling}, while relaxing the linearity assumption most such methods share. Activations are embedded with UMAP, clustered with HDBSCAN, and represented by fuzzy token-to-concept memberships. These memberships enable Concept-Based Alignment (CBA), a fuzzy generalization of the Rand index that compares the pairwise relations induced by two independently discovered concept spaces — without requiring the same architecture, tokenizer, or training stage on both sides, unlike global alignment metrics that presume a shared coordinate system. While the discovered concepts could also support other uses, such as mechanistic interpretability, we focus entirely on their use for alignment, and present a comprehensive view of intra- and inter-model representational alignment across a variety of LLMs.

\begin{figure*}[hbt]
  \centering
  \includegraphics[width=.75\textwidth]{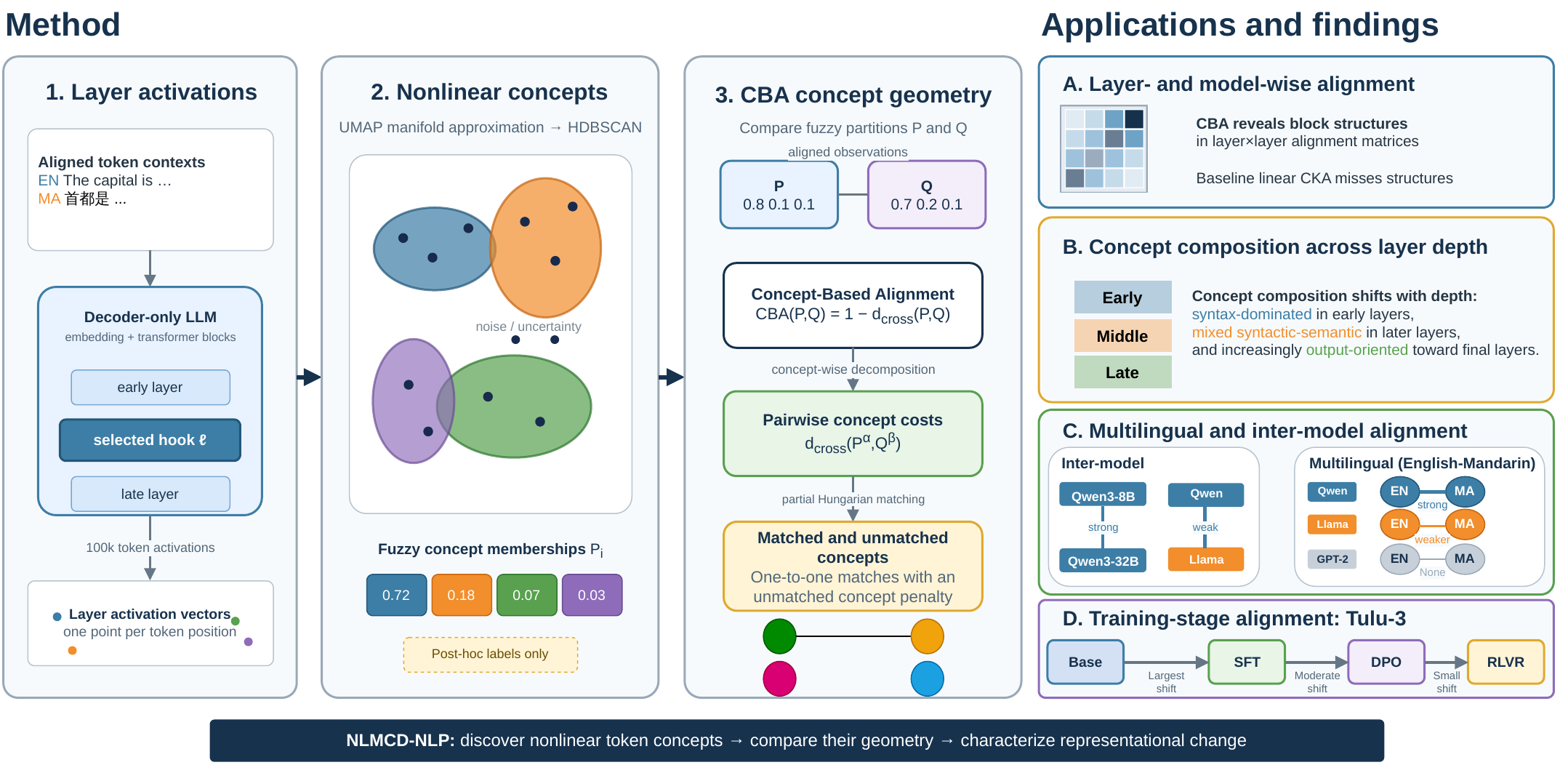}
\caption{Overview of NLMCD-NLP. Layer-wise token activations are clustered into nonlinear concepts with fuzzy memberships. Using a concept-based alignment (CBA) score, we conduct several experiments on these concept manifolds.}
  \label{fig:pipeline}
\end{figure*}

Our technical contributions and most important findings are:
(R0) We adapt NLMCD to residual stream hidden activations and validate the resulting clustering via coverage, reconstruction, and density-based quality metrics. (R1) We propose a neighboring-layer sanity check showing CBA is more sensitive than PCA- and CKA-based baselines. (R2) We find qualitatively similar block structures in layer$\times$layer alignment across models, with two characteristic blocks in intermediate and late layers. (R3) We characterize discovered concepts quantitatively in terms of syntactic/semantic content, finding a relative decrease of purely syntactic concepts in favor of joint syntactic-semantic concepts toward later layers, alongside increasing output-orientation toward final layers. (R4) We find multilingual concepts shared across an aligned English–Mandarin corpus, strongest in Qwen, weaker in Llama, and absent in GPT-2. (R5) We find strong inter-model alignment between same-family Qwen models of different scale, consistent with the block structures identified in (R2), but weak structural alignment across model families. (R6) We trace representational alignment across Tulu-3 training stages: alignment is highest between adjacent stages, validating CBA; the largest shift occurs between the base model and SFT; and preference-alignment (DPO, RLVR) leaves early layers largely unchanged while RLVR mostly preserves DPO's concepts in late layers. Fig.~\ref{fig:pipeline} summarizes the method and the three main uses studied in this paper.

\section{Non-linear concept discovery}
\label{sec:method}

\paragraph{Activation extraction.}
For every model and language, we process 5,000 sentences in batches of eight, truncate them to 128 tokens, and retain a shared reservoir of 100,000 token positions using reservoir sampling \citep{Vitter}. A common sampling stream ensures that row $i$ denotes the same token position at every layer. We record the residual output of each of the transformer blocks for all layers of the respective model. We look at models with a hidden dimension ranging from 768--5,120 across 12--64 layers. Residual activations are L2 normalized before concept discovery. 

\paragraph{UMAP, HDBSCAN, and fuzzy memberships.}
We follow the concept discovery approach of \citet{vielhaben2026beyond} even including hyperparameter settings. For each layer independently, UMAP maps the activation reservoir to 50 dimensions with 30 neighbors, minimum distance 0.01, and seed 42 \citep{mcinnes2018umap}. HDBSCAN then discovers density-supported clusters with minimum cluster size and minimum samples both set to 50, leaf cluster selection, and explicit noise assignments \citep{campello2013hdbscan}. We use one fixed profile across layers and checkpoints so that comparative results are not confounded by per-layer hyperparameter optimization. Let $P\in[0,1]^{N\times C}$ be the resulting proximity matrix. $P_{ic}$ is token $i$'s soft membership in concept $c$. Rows are not forced to sum to one. Missing mass represents HDBSCAN noise or uncertainty.

\paragraph{Concept-Based Alignment.}
CBA compares two fuzzy partitions $P$ and $Q$ defined on aligned observations \citep{hullermeier2012fuzzy,vielhaben2026beyond}. For two fuzzy clusterings $P$ and $Q$ and feature vectors $(\phi_i)_{i=1\ldots N}$, define the membership similarity
  $d_{ms}(P(\phi_i),P(\phi_j))=1-\frac{1}{2}\lVert P(\phi_i)-P(\phi_j)\rVert_1$.
The \emph{cross-distance} is defined by
\begin{align}
  d_{\mathrm{cross}}(P,Q)
  &=\frac{2}{N(N-1)}\sum_{i=1}^N\sum_{j=i+1}^N|d_{ms}(P(\phi_i),P(\phi_j))-d_{ms}(Q(\phi_i),Q(\phi_j))|.
\end{align}
We define the \emph{Concept-based alignment (CBA)} score as   $\operatorname{CBA}(P,Q)=1-d_{\mathrm{cross}}(P,Q)$. CBA equals one when both spaces induce identical pairwise membership relations and reduces to the Hüllermeier Rand index for the case of crisp concept assignments. As in \citep{vielhaben2026beyond}, we also define an \emph{alignment score between two clusters} $P^\alpha$ and $Q^\beta$ from two clusterings $P$ and $Q$ via 
\begin{align}
  d_{\mathrm{cross}}(P^\alpha,Q^\beta)
  &=\frac{1}{N(N-1)}\sum_{i=1}^N\sum_{j=i+1}^N||P^\alpha(\phi_i)-P^\alpha(\phi_j)|-|Q^\beta(\phi_i)-Q^\beta(\phi_j)||\,.
\end{align}

\paragraph{Selected model families} We evaluate nine checkpoints from four families: GPT-2 \citep{radford2019language}, our MI baseline for comparison with prior SAE/NMF work and our English-only contrast for the multilingual analysis (Result~4); Llama-3.1-8B-Instruct \citep{grattafiori2024llama3herdmodels}, our main cross-architecture reference (Results~2, 3, 5), kept over Llama~4 since the latter is multimodal MoE; and Qwen3-4B/8B/32B \citep{yang2025qwen3technicalreport}, providing a scale sweep (Result~2) and multilingual coverage (Result~4). Qwen3-8B and Llama-3.1-8B-Instruct form our matched-scale cross-architecture pair (Results~2--4), chosen for their contrasting data provenance and tokenizers. We use dense Qwen3 rather than Qwen3.5/3.6, which are natively multimodal and would confound scale/language effects with training-paradigm differences; Qwen3 remains the most recent text-only checkpoint in this family, a tradeoff we make explicit.

\paragraph{Finetuning case study} For the finetuning-drift case study (Result~6), we use the Tulu-3-8B suite (base, SFT, DPO, final) \citep{lambert2025tulu3pushingfrontiers}, built on Llama-3.1-8B and, unlike most released models, exposing every intermediate training stage. This lets us separate drift due to supervised finetuning from drift due to preference/RL optimization (DPO, then RLVR), and its shared initialization with our Llama-3.1-8B checkpoint additionally enables a secondary comparison against Meta's own RLHF-tuned Instruct release. More details on the desiderata for model choice can be found in App.~\ref{app:model-selection}.

\paragraph{Experimental setup}
Concept discovery uses 5,000 sentences per language. English is sampled from WikiText-103; the joint English--Mandarin discovery corpus is described in Result 4. Coverage is the non-noise token fraction, and Density-Based Clustering Validation (DBCV) assesses cluster separation \citep{doi:10.1137/1.9781611973440.96}. Embedding diagnostics include raw NRMSE over 20,000 sampled pairs. For the R1/R2 within-model comparisons, CBA, PCA-512, linear CKA, and RBF CKA use the same deterministic sample of 5,000 token rows across layers. PCA-512 treats positive principal-component scores as soft concepts.

\section{Related work}
\paragraph{Alignment measures}
Representational alignment measures have been  categorized in recent reviews \citep{sucholutsky2025getting,klabunde2025similarity}. The most widely used alignment measure for neural representations is centered kernel alignment (CKA) \citep{kornblith2019similarity}, which is based on the Hilbert-Schmidt Independence criterion \citep{gretton2005measuring}. It is invariant to orthogonal transformations and isotropic scaling. Often linear kernels are used in CKA for computational efficiency and similar performance compared to non-linear kernels. In vision, NLMCD \citep{vielhaben2026beyond} has been proposed as a way to capture the more finegrained alignment scores that incorporate the finestructure. The score is based on a generalized Rand-index between fuzzy clusterings. This work explores the use of the NLMCD alignment measure in the context of LLMs.

\paragraph{Zero-shot representation alignment} A separate line of work addresses cross-space alignment without assuming explicit feature correspondence or a shared coordinate system. \citet{moschella2023relative} propose relative representations, enabling zero-shot stitching between independently trained latent spaces. \citet{alvarez2018gromov} instead frame cross-lingual word embedding alignment as an unsupervised Gromov-Wasserstein optimal transport problem. Both approaches share our goal of comparing representation spaces without explicit matching, but operate at a coarser granularity: they align whole embedding spaces or point clouds, whereas CBA compares discovered concept manifolds using alignment scores to be attributed to specific, interpretable concepts rather than reported as a single global statistic.

\paragraph{Concept discovery}
The NLMCD hinges on representational substructures, which are often referred to as concepts. The most commonly considered structures are linear subspaces, in the case of LLMs typically discovered by means of sparse autoencoders with a linear decoder model. This generalized concept definition aligns with recent findings on non-linear substructures in LLM representation spaces \citep{levy2024language,geiger2026world,wurgaft2026manifold,wurgaft2026shape}. Instead of combining SAE directions, this work takes a more direct path and leverages NLMCD's concept definition as general manifolds.

\paragraph{Representational alignment in LLMs}
Cross-architectural alignment has been investigated in \citep{piepereit2026architecture} revealing that trained models show above-random representational alignment, which is  weak in absolute terms and strongly architecture-dependent. It has also been used in \citep{potter2026representational} to assess representational similarity in multi-agent interactions. \citet{shen2026efficient} investigated latent reasoning using CKA. CKA-based alignment scores have also been used for specific applications, such as REEF \citep{ICLR2025_77e830bd} and AWM \citep{zeng2026awm} to derive representation-based fingerprints for LLMs. Overall, CKA has been used to assess representational alignment within and across LLMs, but CKA as alignment measure has rarely been contested. Second, there is not a dedicated alignment study that investigates representational similarity intra- and inter-model in a comprehensive fashion. This is the gap that this submission aims to fill.

\section{Results}

\paragraph{Result 0: Concept Quality Across Layers} 
\label{sec:result0}

\begin{figure}[ht]
  \centering
  \includegraphics[width=\textwidth]{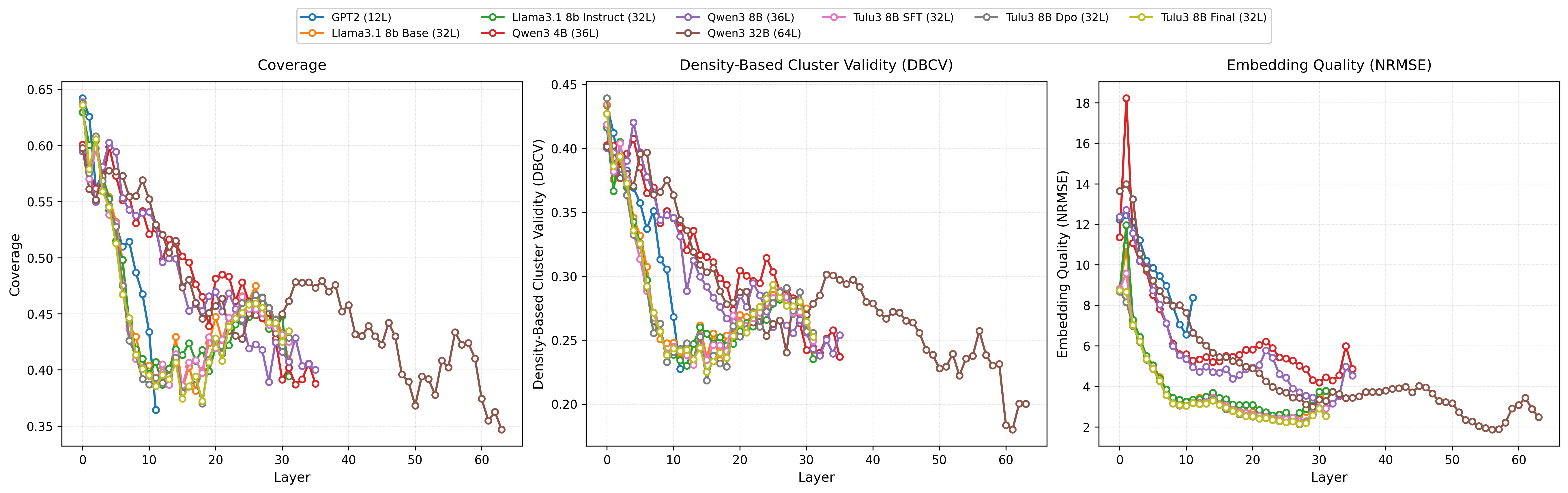}
  \caption{Concept Quality Across Layers: The quality of the concepts extracted from each model is shown across the network depth. For each model, concepts are extracted and evaluated at every layer. (Left) Coverage: the fraction of non-noise tokens. (Middle) Density-based clustering validation (DBCV), which assesses cluster separation. (Right) Embedding quality (NRMSE): the raw normalized root-mean-square error between pairwise distances in the activation space and the reduced embedding.}
  \label{fig:concept_quality_overview}
\end{figure}

Across models, we observe a median coverage, i.e., fraction of non-noise tokens,  of 0.4469, 
substantially higher than the $\sim$0.2 typical of vision foundation models \citep{vielhaben2026beyond}, likely reflecting the greater semantic richness of language relative to vision. Coverage declines from $\sim$0.65 in early layers to $\sim$0.4 in later layers across all models, consistent with the increasing difficulty of capturing the richer representational structure of higher-level features; even so, late-layer coverage remains substantial by vision-domain standards. DBCV, a density-based validity index contrasting intra- vs.\ inter-cluster density, bounded in $[-1,1]$ with higher indicating better-separated clusters, shows a median of 0.2743, 
somewhat lower than the $\sim$0.5 typical in vision foundation models, and correlates strongly with coverage, again pointing to superior clustering quality in earlier layers. Reconstruction quality (NRMSE, see App. \ref{app:embedding-distance}) follows the same trend, falling from 10-12 in early layers to 3-4 in later layers, converging to values comparable to those reported in the vision setting. Taken together, these results support the technical validity of the concept discovery step itself. We highlight the robustness of the NLMCD approach which transfers from vision to LLMs without the necessity to adapt hyperparameter settings.

\paragraph{Result 1: Sanity check}
As a sanity check for the soundness of our alignment measure, we exploit the assumption that adjacent layers should show the strongest representational correspondence: for each source layer, we ask whether its best-aligned target layer under a given alignment method, excluding self-alignment, is a direct neighbor. The \textit{NN alignment score} reports the fraction of layers for which this holds; \textit{Neighbor z} additionally reports the maximum alignment score (z-standardized across rows) of the direct neighbors to rule out uninformative flat alignment profiles. We compare CBA to CKA with RBF kernel as non-linear baseline and two linear baselines, linear CKA and PCA with 512 dimensions, averaged over 9 model checkpoints (Tab.~\ref{tab:nn-alignment-margin}; per-model results in App. Tab.~\ref{tab:nn-alignment-margin-per-model}).
All four methods show nearly perfect NN alignment scores, with slight advantages for the non-linear alignment measures CKA(RBF) and CBA, with 0/1 mispredicted neighbors across all layers and models. The Neighbor z score represents the more informative metric in this case and clearly identifies CBA as most informative measure. While the two non-linear methods perform on par in terms of NN Alignment Score, CBA enjoys the additional benefit that it yields concepts as substructures and allows for a more finegrained characterization of the alignment between these substructures.

\begin{table}[ht]
\centering
\begin{tabular}{lcc}
\toprule
Method & NN Alignment Score $\uparrow$  & Neighbor $z$ $\uparrow$ \\
\midrule
PCA (512-d) & 0.983 $\pm$ 0.025  & 1.867 $\pm$ 0.244 \\
CKA (linear) & 0.978 $\pm$ 0.029 & 1.287 $\pm$ 0.486 \\
CKA (RBF) & \textbf{1.000 $\pm$ 0.000} & 1.714 $\pm$ 0.182 \\
CBA (NLMCD) & 0.997 $\pm$ 0.009 &  \textbf{2.515 $\pm$ 0.256} \\
\bottomrule
\end{tabular}
\caption{Nearest-neighbor alignment score and best-neighbor z-score averaged equally over 9 model checkpoints (mean $\pm$ population standard deviation). All four methods recover adjacent-layer correspondence almost perfectly, so the NN alignment score alone provides limited discrimination between them. Neighbor $z$ measures the strongest neighbor relative to the row mean and standard deviation. Per-model results are reported in App. Tab.~\ref{tab:nn-alignment-margin-per-model}.}
\label{tab:nn-alignment-margin}
\end{table}

\begin{figure*}[!t]
  \centering
  \includegraphics[width=\textwidth]{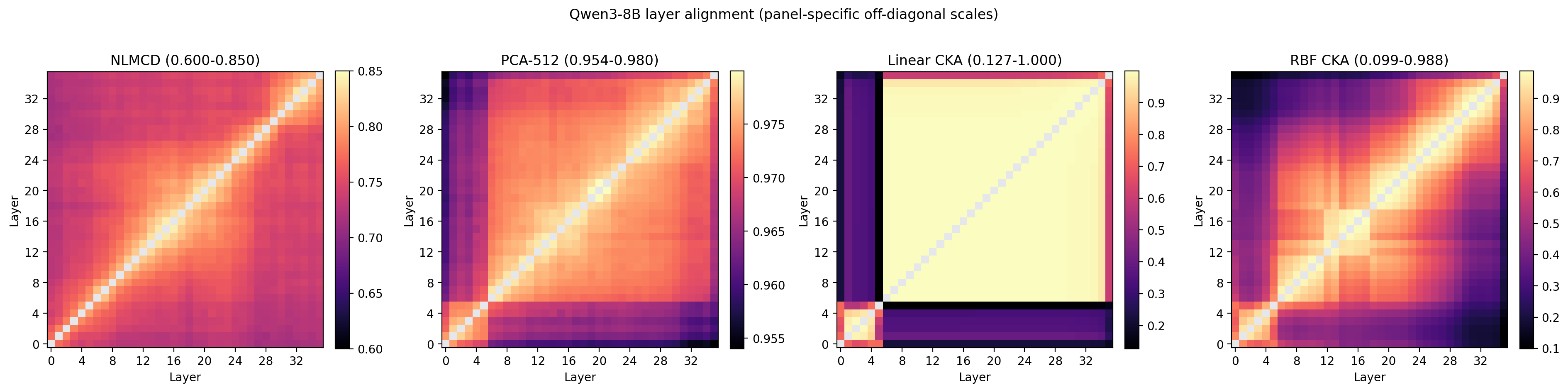}
\caption{Qualitative comparison of alignment plots for Qwen3 8B}
  \label{fig:qwen3_8b_nlmcd_pca_cka}
\end{figure*}

We also compare the expressiveness of each metric choice in Fig.~\ref{fig:qwen3_8b_nlmcd_pca_cka}. When using linear CKA as the alignment metric we can identify two clear block structures but all finer details between layers 6-36 are not visible. In contrast both the PCA-512 and NLMCD methods reveal structures that linear CKA misses. RBF CKA also reduces the near-saturation seen with linear CKA. This stresses that the choice of the alignment metric strongly affects the observable substructures within the model.

\begin{figure}[htbp]
\centering
\setlength{\fboxsep}{0pt}
\setlength{\tabcolsep}{0pt}

\begin{subfigure}[b]{0.23\textwidth}
  \centering
  \includegraphics[width=\linewidth]{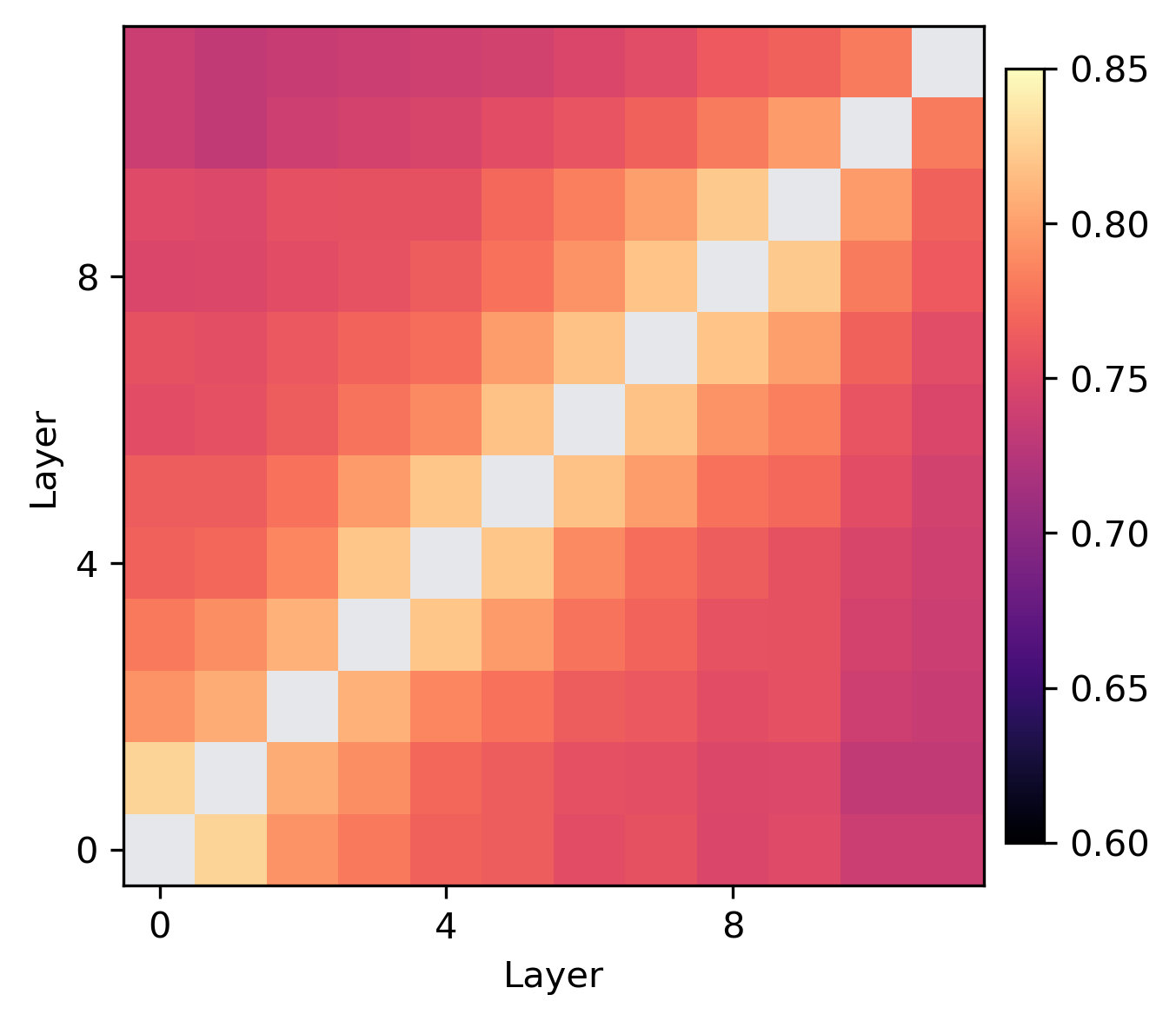}
  \caption{GPT2}
  \label{fig:1}
\end{subfigure}%
\hfill
\begin{subfigure}[b]{0.23\textwidth}
  \centering
  \includegraphics[width=\linewidth]{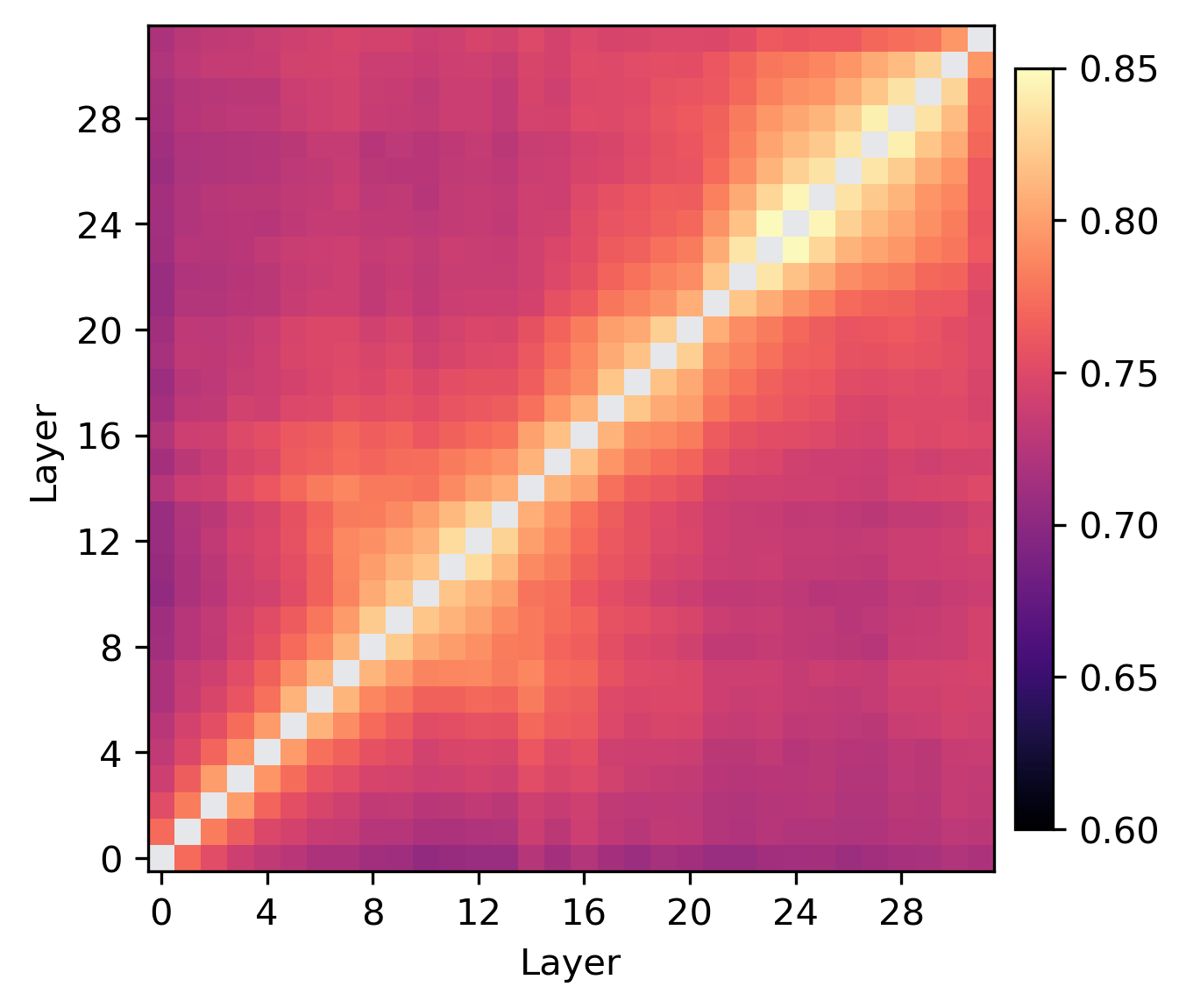}
  \caption{Llama3.1 8B Instruct}
  \label{fig:3}
\end{subfigure}
\hfill
\begin{subfigure}[b]{0.23\textwidth}
  \centering
  \includegraphics[width=\linewidth]{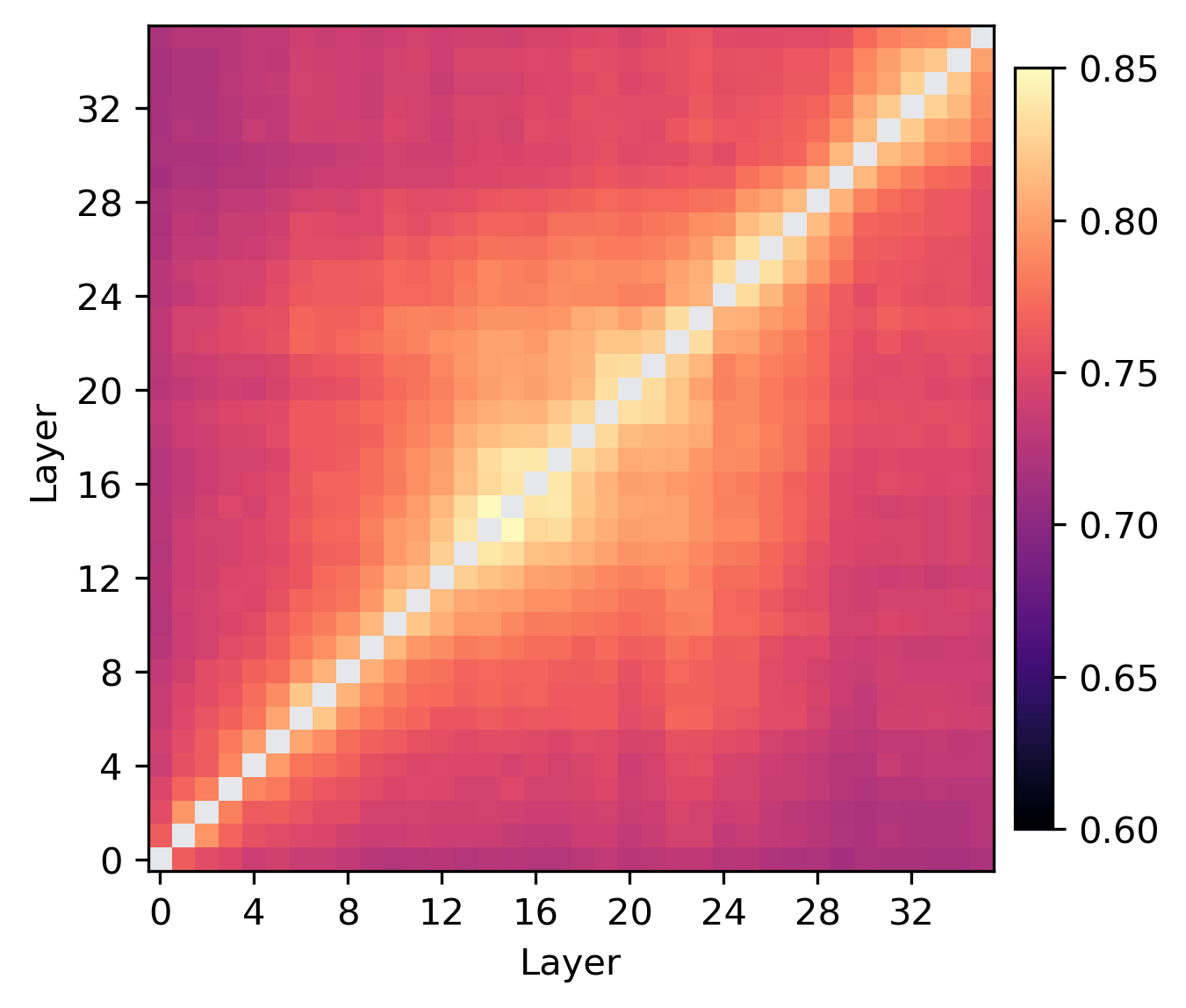}
  \caption{Qwen3 4B}
  \label{fig:4}
\end{subfigure}
\hfill
\begin{subfigure}[b]{0.23\textwidth}
  \centering
  \includegraphics[width=\linewidth]{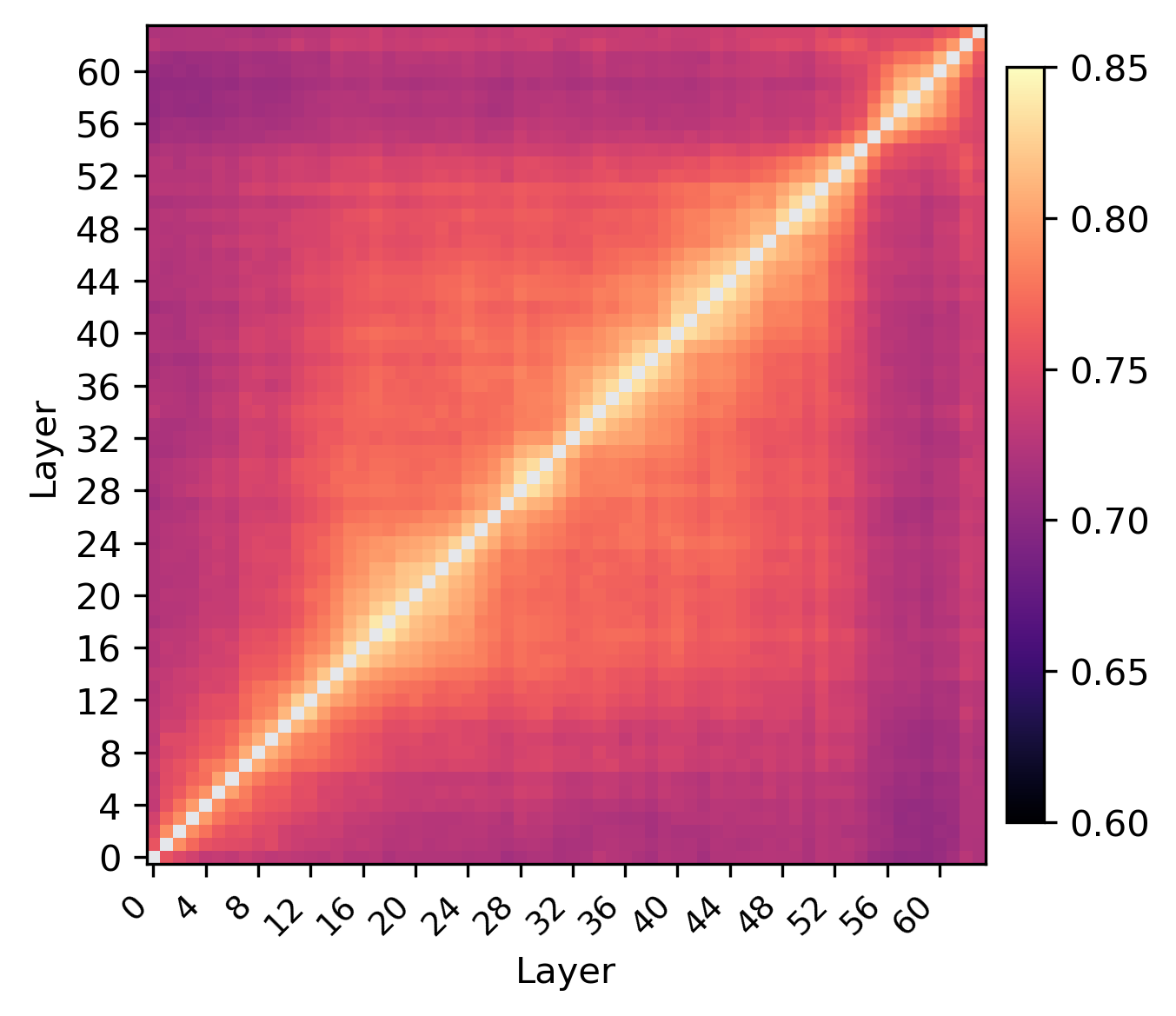}
  \caption{Qwen3 32B}
  \label{fig:6}
\end{subfigure}

\caption{CBA layer alignment. Shared color scale $[0.60,0.85]$; self-alignment diagonals masked. Layer indices are zero-based. All panels use 5,000 samples. Refer to the first subplot in Fig.~\ref{fig:qwen3_8b_nlmcd_pca_cka} for the Qwen3 8B result and to Fig.~\ref{fig:cba_grid_app} for an extended plot comparing all models.}
\label{fig:cba_grid}
\end{figure}

\paragraph{Result 2: Layer-wise alignment} 
Fig.~\ref{fig:cba_grid} shows the CBA layer-alignment matrices for four models spanning different model families and parameter counts. All matrices exhibit a pronounced diagonal structure, with model-specific variations in the arrangement of aligned layers. For the larger-scale models, except for GPT, the matrices display a band-diagonal structure in the first region, followed by a block structure, a second band-diagonal region, and a final block structure. These patterns are qualitatively similar across the models. All Qwen models share a larger block structure in their middle layers. Compared with the 8B models, the 32B models exhibit a more finegrained structure in the central block.  Overall, several larger models exhibit a similar sequence of band-diagonal and block-structured regions, suggesting recurring alignment patterns across model families, see Result 5 for details.

\paragraph{Result 3: Descriptive/qualitative insights} 
We compare properties of annotated concepts aggregated over all residual layers for Llama-3.1-8B-Instruct and Qwen3-8B. Those models were chosen as they are comparable in parameter count and number of layers. Concepts were annotated by a labeling LLM along two axes: concept type (syntactic, semantic, mixed, or unclear) and evidence orientation (input, output, both, or unclear). Orientation describes which token and next-token evidence best explains the concept. Semantic and syntactic concept percentages are roughly on-par in Llama (45.7\% and 45.1\%), while syntactic concepts dominate in Qwen (53.2\% versus 39.8\% semantic). Regarding the orientation annotation results, input-oriented concepts clearly dominate in both models (93.4\% and 96.3\% respectively), while concepts with both input and output evidence account for 6.6\% and 3.7\%. Output-oriented concepts are omitted from the table as neither model has any, see Tab.~\ref{tab:r3-concept-annotations} for details.
\begin{figure*}[!t]
  \centering
  \begin{subfigure}[t]{0.38\textwidth}
    \includegraphics[width=\linewidth]{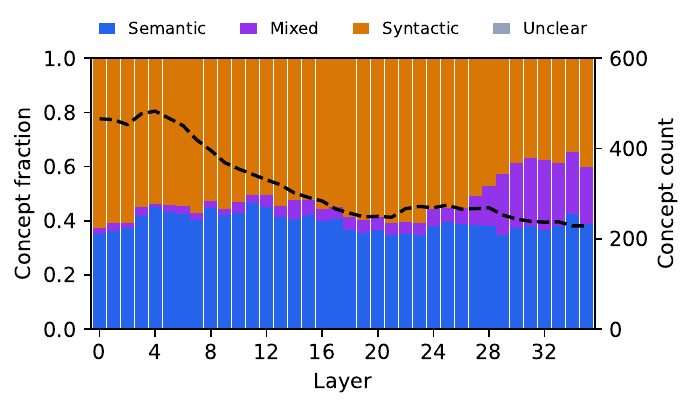}
    \caption{Qwen3-8B: semantic/syntactic}
  \end{subfigure}\hspace{0.025\textwidth}
  \begin{subfigure}[t]{0.38\textwidth}
    \includegraphics[width=\linewidth]{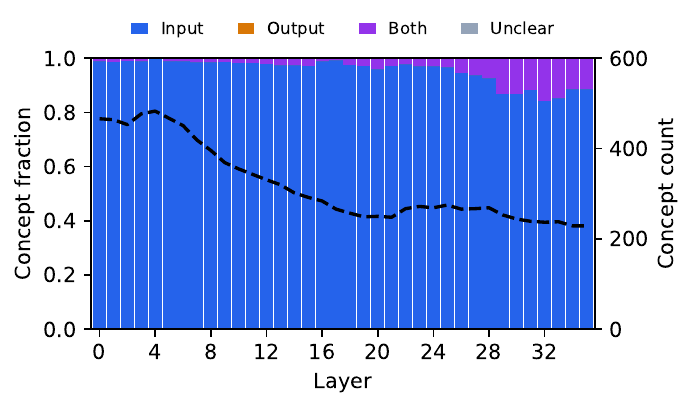}
    \caption{Qwen3-8B: input/output}
  \end{subfigure}
  \caption{English concept categories and evidence orientation over residual depth. Bars show concept fractions (left axis). The dashed black line shows the total number of annotated concepts at each layer (right axis). Analogous plots for Llama 3.1 8B are shown in Fig.~\ref{fig:r3-app}.}
  \label{fig:r3-concept-types}
  \label{fig:r3-concept-orientation}
\end{figure*}
Fig.~\ref{fig:r3-concept-orientation} and App.~Fig.~\ref{fig:r3-app} summarize the layer-dependence of both quantities. The fraction of syntactic concepts stays relatively high across the first two thirds (Qwen)/the first half (Llama) of the model and then shrinks as the fraction of jointly syntactic and semantic concepts starts to rise. This is partly consistent with probing studies on BERT, which found that syntactic information is more locally encoded in earlier layers while semantic tasks benefit from later-layer representations \citep{tenney2019bertrediscoversclassicalnlp}. Our unsupervised concepts show the syntactic-dominance side of this pattern, but show a more nuanced picture, with joint semantic/syntactic concepts becoming more prevalent in later layers. For orientation-annotated concepts the fraction of output-oriented concepts stays close to 0\% across all layers. There is, however, a clear increase of concepts annotated as having both input and output evidence towards the later layers in both models.

\paragraph{Result 4: Multilingual activation} 
In order to quantify the amount of multi-lingual activation taking place within the selected models, we measure joint parallel activation on a parallel corpus of English and Mandarin sentences. To do this, we first conduct concept discovery on a separate joint English-Mandarin text corpus based on English WikiText records and Mandarin Wikipedia sentences. 5,000 sentences are taken in streaming order after filtering from each corpus for concept discovery. 
After this a deterministic random 5,000-pair sample of Tatoeba English-Mandarin translations is used as a parallel sentences probe. Concept assignments are computed on the joint clusters for each token in both sentences. Memberships are summed over tokens within each sentence and normalized across concepts to obtain distributions $p_{\mathrm{en}}$ and $p_{\mathrm{zh}}$. Shared mass is $\sum_c \min(p_{\mathrm{en},c},p_{\mathrm{zh},c})$, averaged over valid translation pairs.

\begin{table*}[!ht]
\centering
\small
\begin{tabular}{lrrrrr}
\toprule
Model & Paired mass & Shuffled mass & Mean excess & Peak layer & Peak excess \\
\midrule
GPT-2                    & 0.141 & 0.140 & 0.0009 & 5  & 0.0039 \\
Llama-3.1-8B-Instruct    & 0.176 & 0.152 & 0.0242 & 2  & 0.0980 \\
Qwen3-4B                 & 0.245 & 0.196 & 0.0487 & 7  & 0.1187 \\
Qwen3-8B                 & 0.244 & 0.201 & 0.0430 & 8  & 0.1012 \\
Qwen3-32B                & 0.286 & 0.247 & 0.0392 & 10 & 0.0817 \\
\bottomrule
\end{tabular}
\caption{English--Mandarin shared concept mass and shuffled-pair controls over residual layers; peak layer (zero-based) indicates the layer with the largest excess with excess indicated in peak excess.}
\label{tab:r4-parallel-activation}
\end{table*}

Tab.~\ref{tab:r4-parallel-activation} shows the results of the multilingual activation experiment. Paired mass is the raw shared mass computed from the parallel sentences probe. However, especially in the larger models we observed layer-specific peaks that did not seem to correlate with multilingual activation, see App. Fig.~\ref{fig:r4-parallel-activation}. We therefore computed a baseline using 200 permutations of the Mandarin sentences per layer, recomputing shared mass and averaging over permutations (shuffled mass). Mean excess is paired mass minus shuffled mass, averaged equally over residual layers. Paired mass and shuffled mass in the table are also layer averages.

In line with our expectation the excess shared mass is close to zero for GPT-2, our predominantly English-trained comparison model and increases for the multilingual Llama and Qwen models. The mean excess increases from 0.0242 to 0.0487 from Llama to Qwen3-4B. After that there is a slight drop of mean excess for Qwen3-8B (0.0430) and Qwen3-32B (0.0392). The peak occurs at layer 7 for Qwen3-4B, layer 8 for Qwen3-8B, and layer 10 for Qwen3-32B. The finding that multilingual concept sharing between typologically distant languages is highly training-dependent rather than a universal property provides an empirical counterpoint to the Platonic Representation Hypothesis \citep{huh2024platonicrepresentationhypothesis}, in line with prior work on linear SAE representations \citep{brinkmann-etal-2025-large}.

\begin{figure*}[!ht]
  \centering
  \includegraphics[width=.9\textwidth]{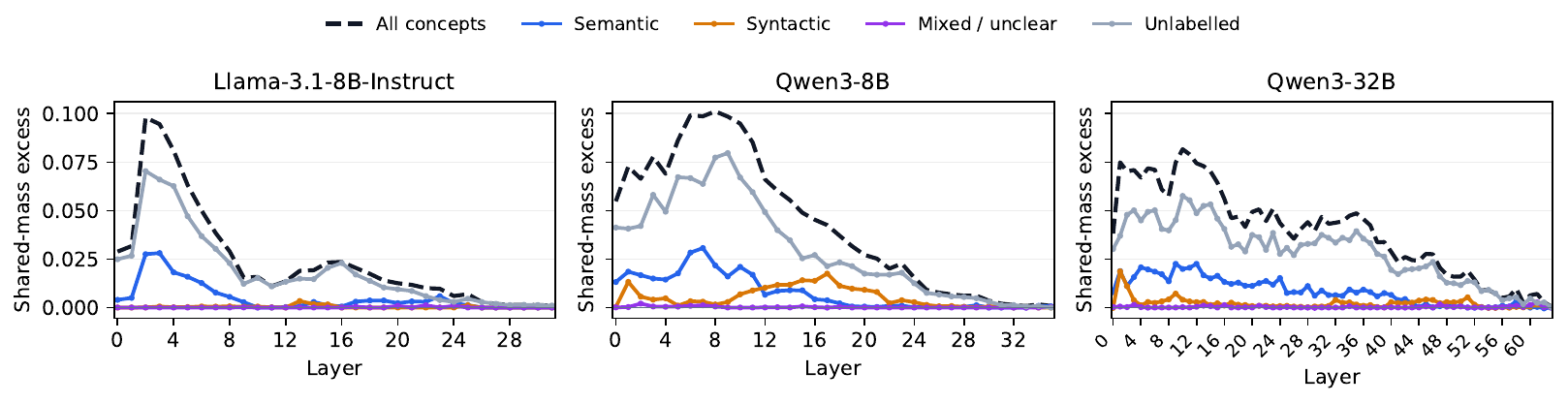}
  \caption{Semantic, syntactic, mixed, and unlabeled shared-concept mass above shuffled-pair controls for Llama-3.1-8B-Instruct, Qwen3-8B, and Qwen3-32B. Negative excess is retained; all panels share the same vertical scale.}
  \label{fig:r4-semantic-syntactic-mass}
\end{figure*}

In addition to reporting the mean excess for shared mass across layers we also annotated multilingual concepts using a labeling LLM, labeling them as either semantic, syntactic or mixed. At each layer, a concept is selected as translation-shared if active in both sentences of at least 2\% of valid translation pairs and at least five pairs. These are 100 pairs with our 5,000 valid pairs. Each sentence independently satisfies the active criterion through top-10 membership or a normalized concept score $\geq 0.03$. The labeler uses bilingual training tokens and representative parallel examples. Semantic denotes shared meaning, syntactic includes grammar and surface form, mixed requires both, and unclear denotes insufficient evidence.
Fig.~\ref{fig:r4-semantic-syntactic-mass} shows the results of the multilingual concept annotation. All three models show an early-layer peak in semantic shared-mass excess that gradually decreases towards the middle layers. Qwen3-8B and Qwen3-32B also show a local syntactic peak within their first four layers. Among labeled translation-shared concepts, semantic excess is prominent in early layers. Much of the total excess is unlabeled, so this does not establish semantic dominance across all concepts. Shared punctuation and sentence structure are possible explanations for syntactic contributions.

\paragraph{Result 5: Cross-model alignment} 

\begin{figure}[ht]
  \centering
  \includegraphics[width=0.23\textwidth]{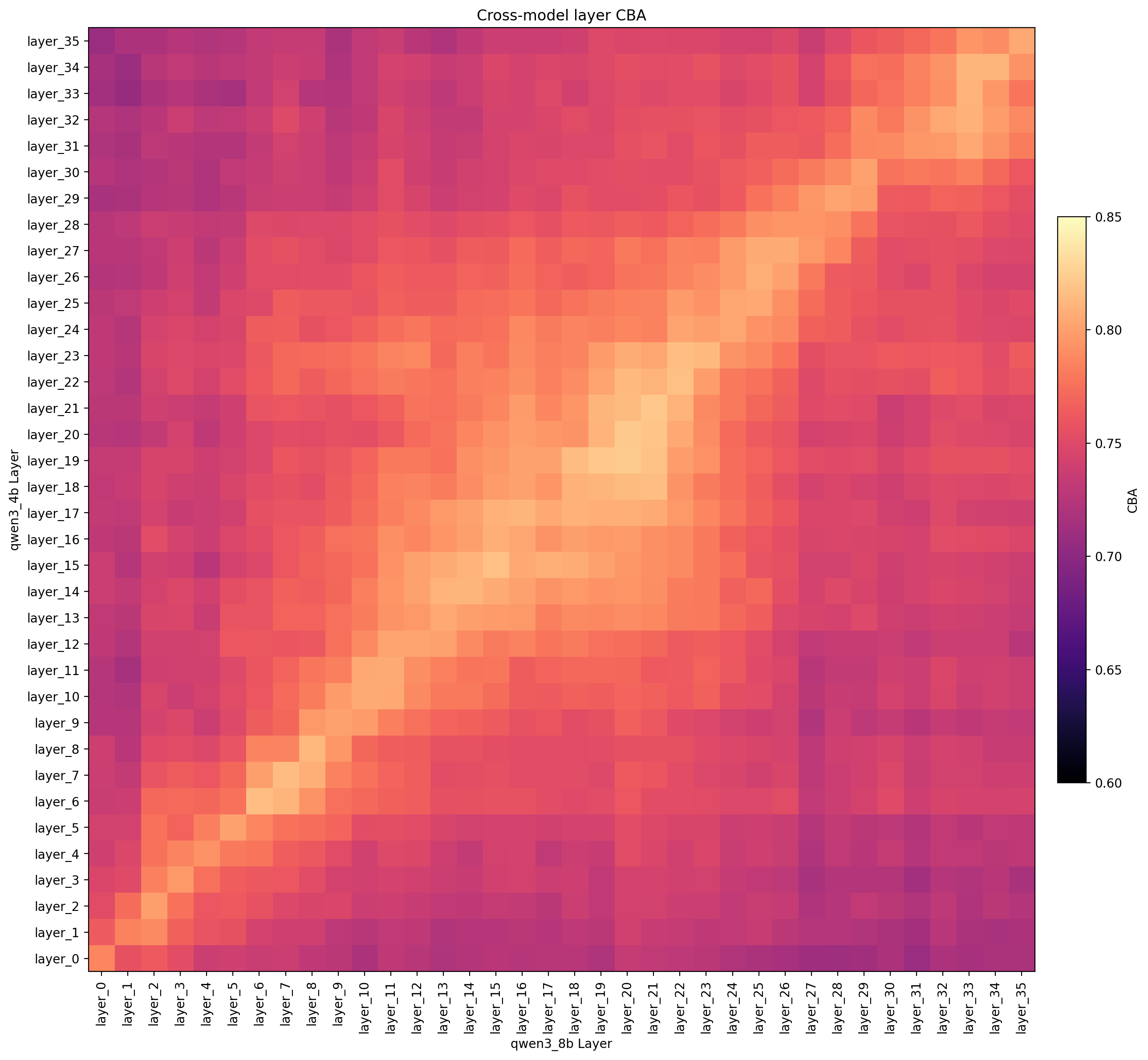}
  \hfill
  \includegraphics[width=0.23\textwidth]{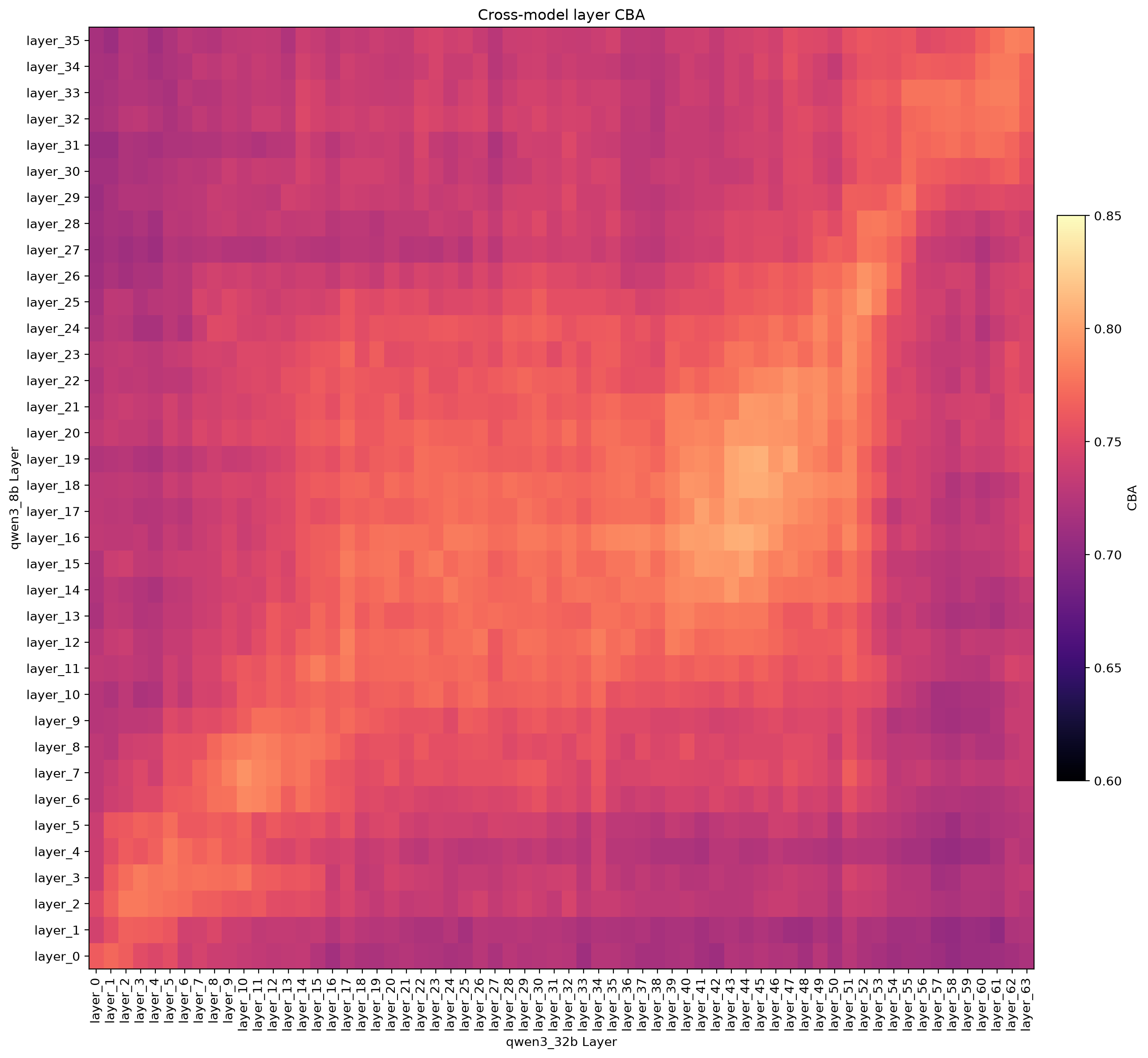}
  \hfill
  \includegraphics[width=0.23\textwidth]{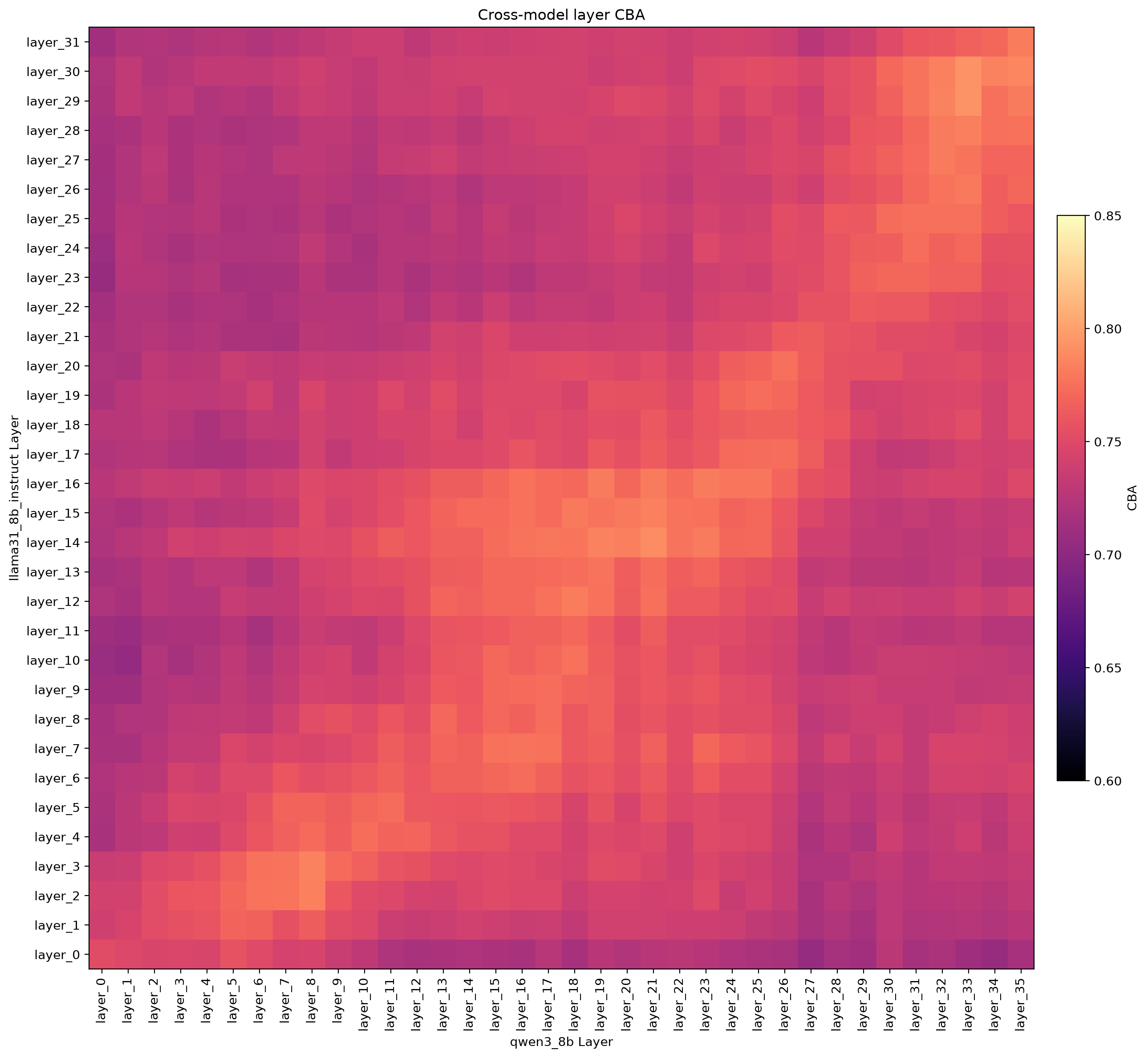}
  \caption{CBA Similarity Scores between models. Left: Qwen 4b vs 8b, Center: Qwen 8b vs. 32b, Right: Llama 8b instruct vs. Qwen 8b}
  \label{fig:CBA_models}
\end{figure}

Fig.~\ref{fig:CBA_models} shows CBA matrices for comparisons between Qwen and Llama model variants. Because Qwen and Llama use different tokenizers, we align their tokens using character spans within the same sentence. We retain a target token only when sampled source-token spans lie wholly within and together cover its complete span. The source tokens' concept memberships are then mean-pooled before comparison. All R5 CBA comparisons use 1,024 aligned observations per model. Both within-family Qwen comparisons exhibit pronounced depth-structured alignment. High CBA scores form a broad diagonal band, indicating that layers at comparable relative depths encode similar concepts. The alignment comprises an early band, a broad middle block, and a smaller high-alignment region near the output layers; the main difference between Qwen~8B and 32B is an extended middle block. This pattern persists despite differences in scale and layer count, suggesting that Qwen variants preserve a common progression of internal representations.
In contrast, Llama~8B Instruct versus Qwen~8B yields substantially lower CBA scores, although a weak diagonal and middle-layer block structure remain. This indicates limited, locally structured cross-family concept alignment, consistent with prior work reporting weak overall cross-family but strong within-family alignment \citep{piepereit2026architecture}. Nevertheless, the result shows that the internal organization does not look similar, see Result 2, but rather supports the hypothesis of corresponding functional blocks even across different model families.

\begin{figure*}[hbt]
  \centering
  \includegraphics[width=0.44\textwidth]{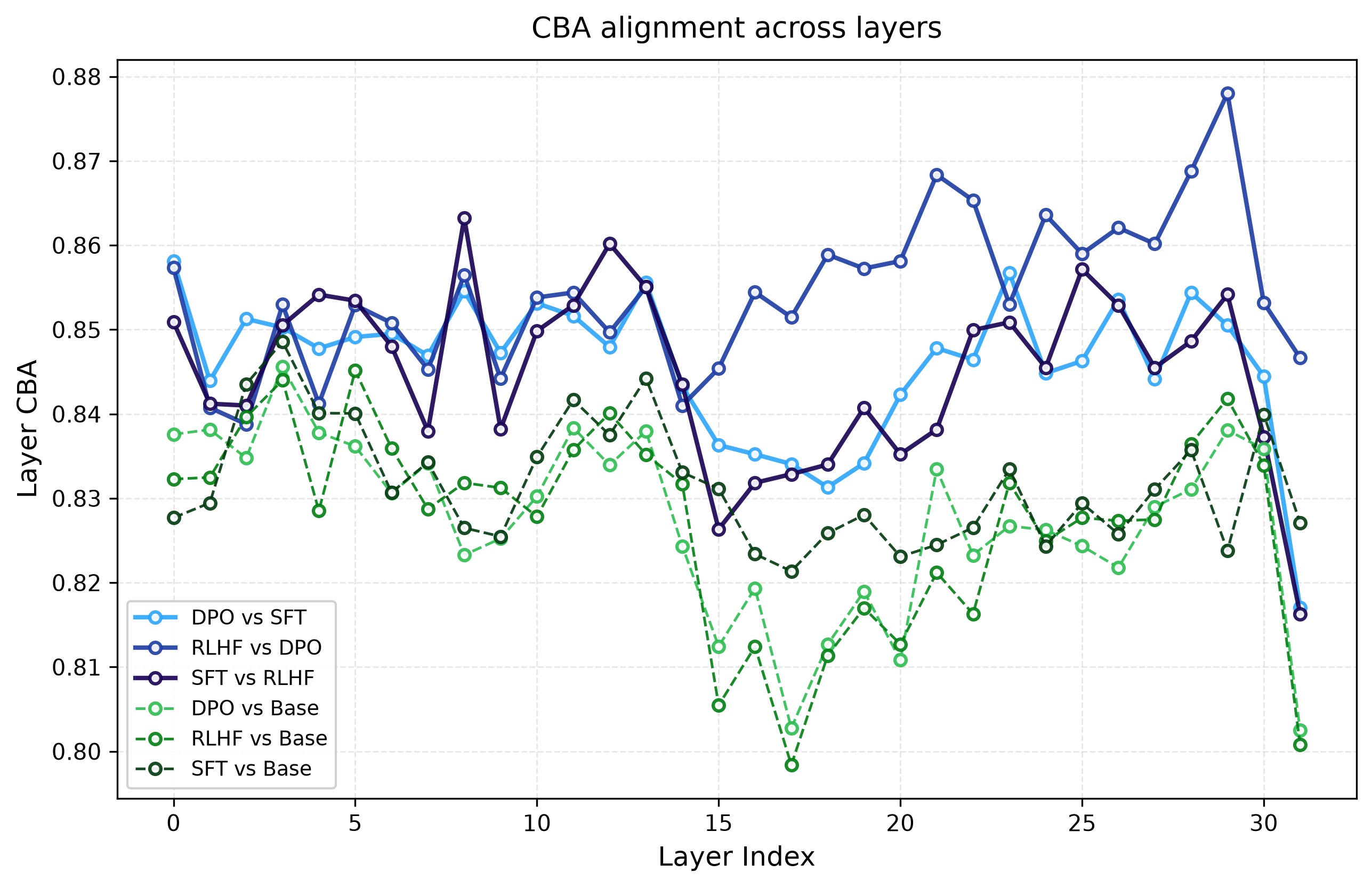}
\caption{Pairwise same-layer CBA alignment across the Llama-3.1-8B base model and the SFT, DPO, and final RLVR checkpoints of Tulu-3.}
  \label{fig:CBA_alignment_finetune}
\end{figure*}

\paragraph{Result 6: Finetuning case study} 
We compare the three finetuned Tulu-3-8B variants with their base model, Llama-3.1-8B. Fig.~\ref{fig:CBA_alignment_finetune} reports CBA scores for corresponding layers across model pairs. Alignment between adjacent training stages is high, consistent with the claim that objective consistency supports consistent representational similarity patterns \citep{pmlr-v267-ciernik25a}. DPO and the final RLVR model are particularly closely aligned, with scores remaining near $0.85$--$0.86$ across most layers, suggesting that RLVR largely preserves the concept geometry established during DPO rather than reorganizing intermediate representations. Comparisons involving the SFT model show lower, intermediate alignment, indicating a larger representational change during supervised finetuning. This is consistent with parameter-level evidence on the same checkpoints: SFT induces dense updates, including from Llama-3.1-8B to Tulu-3-8B-SFT, whereas DPO leaves about 81\% of the SFT model's parameters unchanged \citep{mukherjee2025reinforcementlearningfinetunessmall}. Across all stages, alignment remains above $0.8$, higher than typical cross-model alignment scores, suggesting that most concepts are formed during pretraining and are only moderately modified by subsequent finetuning.

Most model pairs show an alignment drop from layer~14 to layer~15, which is substantially larger in comparisons involving the base model. The effect is not confined to layer~15: alignment remains reduced until approximately layer~22, after which it gradually recovers and, for DPO--RLVR, eventually exceeds its layer-14 value. Alignment also drops sharply at the final layer in every comparison, consistent with increased specialization toward next-token prediction. The same pattern is visible in the alignment matrices in App.~Fig.~\ref{fig:CBA_models_finetune}. Overall, alignment changes are larger relative to the base model than between successive finetuning stages, with DPO and the final RLVR checkpoint remaining most closely aligned.
The fact that across all finetuning stages the drop occurs after layer 14 indicates that this layer is of particular importance for the behavior of the model that is supposedly adapted during post-training.

To investigate the alignment drop between layers~14 and~15, we apply partial concept matching based on the Hungarian algorithm. We augment the concept-cost matrix with dummy assignments whose cost equals the match-acceptance threshold, set to $0.0025$ in all experiments. This allows concepts to remain unmatched when all real assignments exceed that cost, rather than forcing high-cost one-to-one matches. Fig.~\ref{fig:concept_cost_matrix_sft_dpo} overlays the resulting matches on the pairwise cost matrices for SFT and DPO, where layer-15 shows a weaker and less coherent diagonal structure than layer~14.

App.~Tab.~\ref{tab:match_examples_sft_dpo} shows representative SFT--DPO concept matches at layers~14 and~15. Layer-14 matches are generally more semantically coherent, with the highest-scoring pairs consisting predominantly of function- or structure-related tokens. In contrast, layer-15 matches are more heterogeneous and often semantically weaker; even highly ranked pairs can mismatch, e.g.\ ``Name beginnings'' vs. ``Equality symbols''. Layer~15 also yields more unmatched concepts in both directions. Together with the lower CBA alignment, these observations suggest that the layer-14 to 15 transition changes the concepts represented after DPO rather than uniformly reducing similarity.

\section{Limitations and future work}
First, we do not evaluate models above $\sim$32B parameters or mixture-of-experts architectures. This is partly a compute constraint, and partly because our core claims concern the \textit{existence} and \textit{cross-layer structure} of non-linear concept manifolds rather than their scaling behavior at frontier scale. However, we acknowledge both directions as very interesting directions for future research. Second, the computational complexity of the approach is dominated by the HDBSCAN clustering step. Given our time and budget constraints this limits the number of tokens to $O(100k)$. Devising a computationally less expensive alternative is therefore a worthwhile direction for future research. Third, the discovered concepts rely on the extraction corpus. Even though Wikipedia text provides a broad basis, some of the semantic concepts in intermediate layers are highly specific and further ones at a similar level of detail might not have been sufficiently activated. Finally, the approach can be used for more in-depth studies of the identified concepts along the lines sketched in Result 6.

\clearpage
\subsection*{AI use statement}
In this work, we used generative AI tools for tasks that require disclosure under current publication guidelines, specifically to assist with method implementation and to suggest improvements to existing implementations. We have not used generative AI tools for conceptual development, framing of results, formulation of hypotheses, design of methodology or experiments, interpretation of results, or the writing of proofs. Tasks such as generating synthetic datasets, developing theoretical models or conceptual frameworks, formulating mathematical claims, and providing critical ingredients for proving mathematical claims are not applicable to this work.

Additionally, we used generative AI tools for tasks that fall under recommended disclosure, including: (i) creating or editing software code (e.g., drafting boilerplate, refactoring, and suggesting alternative implementations, refinement of plotting routines); (ii) editing the manuscript to improve readability and language quality; (iii) assisting with model selection among models to be evaluated (e.g., summarizing documented capabilities, constraints, and typical use cases); and (iv) supporting routine writing subtasks such as proposing section structure, phrasing alternatives, and formatting references. No AI tool was used to generate scientific figures or images, to transcribe recordings, or to perform literature discovery or screening.

We have reviewed all AI‑assisted work. In particular, LLM‑generated or LLM‑modified code was inspected line‑by‑line, executed in our experimental environment, and validated against unit tests and ablation checks by at least two authors; any suggestions affecting experimental logic were cross‑checked against the original method descriptions and implementation notes. Language edits and structural suggestions were accepted only when they preserved the technical meaning and did not introduce new claims; all such changes were manually verified. We take responsibility for the final content of this work, including text, claims, code, and any other artifacts produced with the aid of generative AI.

\subsection*{Ethics statement}
All authors have read and agree to comply with the ICLR Code of Ethics.

\subsection*{Reproducibility statement}
We make the full source code underlying our submission available in an anonymized code repository \url{https://anonymous.4open.science/r/NLMCD-NLP-C5E7}.

\clearpage
\bibliography{iclr2027_conference}
\bibliographystyle{iclr2027_conference}

\clearpage
\appendix
\section{Appendix}
\subsection{Model Selection}
\label{app:model-selection}

We evaluate our approach on nine checkpoints drawn from four model families, chosen to jointly satisfy four criteria: (1) \textit{architectural purity} --- dense, purely text-trained decoders wherever a claim is language-specific, so that representational geometry is not confounded with joint vision-language training; (2) \textit{comparability to prior mechanistic-interpretability work}, since Result~1 sanity-checks our measure against CKA and PCA baselines reported in the literature; (3) \textit{targeted coverage} of the specific axis each result section requires (sanity-checking, scale, cross-architecture alignment, multilinguality, training stage); and (4) \textit{feasibility under a fixed compute budget}, which rules out broad sweeps across many independent families or model sizes above $\sim$30B parameters. Tab.~\ref{tab:models} summarizes the resulting suite and which result section each checkpoint supports.

\paragraph{GPT-2 \citep{radford2019language}.} Serves as our mechanistic-interpretability anchor. Despite being the oldest model in our suite, it is the model against which the majority of existing SAE- and NMF-based concept-extraction results have been reported, making it essential for situating our manifold-based proximity measure relative to prior linear approaches (Result~1). Its small size and English-only pretraining also make it our contrast case in the multilingual analysis (Result~4).

\paragraph{Llama-3.1-8B-Instruct \citep{grattafiori2024llama3herdmodels}.} Provides our primary cross-architecture reference point at the $\sim$8B scale (Results~2, 3, 5), and supplies the base weights underlying our finetuning-stage case study (see below). We retain this mid-2024 checkpoint rather than substituting a newer Llama release because Llama~4 (April 2025) moved to a mixture-of-experts, natively multimodal early-fusion architecture, which no longer satisfies our architectural-purity criterion; Llama-3.1-8B is thus the most recent Llama checkpoint compatible with our design.

\paragraph{Qwen3-4B, Qwen3-8B and Qwen3-32B \citep{yang2025qwen3technicalreport}.} Supply our within-family scale sweep (Result~2) and our multilingual analysis (Result~4), holding architecture and training recipe fixed while varying parameter count roughly eightfold. Qwen3-8B additionally forms our matched-scale cross-architecture pair with Llama-3.1-8B-Instruct (Results~2--4), chosen as the comparison point because Qwen3 and Llama-3.1 differ substantially in data provenance, tokenizer, and multilingual coverage. We use the dense Qwen3 line rather than the more recent Qwen3.5/3.6 releases because, from Qwen3.5 onward, Alibaba's flagship checkpoints are trained natively multimodally via early fusion of vision and text tokens; using these would confound our scale and multilingual claims with differences in training paradigm rather than isolating parameter count and language as intended. Qwen3 dense models remain purely text-trained, cover 119 languages including Mandarin, and are, to our knowledge, the most recent checkpoints from any major open-weight family that satisfy our text-only constraint at time of writing --- a recency/architectural-purity tradeoff we make explicit rather than an unexamined limitation.

\paragraph{Tulu-3-8B suite: base, SFT, DPO, final \citep{lambert2025tulu3pushingfrontiers}.} Supports our finetuning-drift case study (Result~6). Unlike most publicly released instruction-tuned models, which expose only base and final checkpoints, the Tulu~3 pipeline --- built on Llama-3.1-8B base weights --- publishes each intermediate training stage explicitly. This lets us attribute representational drift separately to supervised finetuning and to preference/RL-based optimization (DPO, then RLVR), rather than collapsing both into a single base-to-instruct delta, which a single additional checkpoint could not support. We chose this suite over comparable alternatives (e.g., OLMo~2) because it shares its base initialization with the Llama-3.1-8B checkpoint used elsewhere in our suite, allowing a direct comparison between Meta's RLHF-tuned Instruct release and AI2's independently-tuned Tulu~3 endpoint as a secondary check on how sensitive our drift measure is to the specific finetuning pipeline, not only its stages. We summarize the selected models in Tab.~\ref{tab:models}.

\begin{table}[!hbt]
\centering
\small
\begin{tabular}{lll l}
\toprule
\textbf{Model}  & \textbf{Release} & \textbf{HF identifier} & \textbf{Result section(s)} \\
\midrule
GPT-2 (124M)                          & 02/2019  & \texttt{openai-community/gpt2} & 0, 1, 4 \\
Llama-3.1-8B-Instruct                 & 07/2024  & \texttt{meta-llama/Llama-3.1-8B-Instruct} & 0, 1, 2, 3, 4, 5 \\
Qwen3-4B                               & 05/2025  & \texttt{Qwen/Qwen3-4B} & 0, 1, 2, 4, 5 \\
Qwen3-8B                               & 05/2025  & \texttt{Qwen/Qwen3-8B} & 0, 1, 2, 3, 4, 5 \\
Qwen3-32B                             & 05/2025  & \texttt{Qwen/Qwen3-32B} & 0, 1, 2, 4 \\
Llama-3.1-8B (base)                       & 06/2024  & \texttt{meta-llama/Llama-3.1-8B} & 1, 6 \\
Tulu-3-8B (SFT)                        & 11/2024  & \texttt{allenai/Llama-3.1-Tulu-3-8B-SFT} & 1, 6 \\
Tulu-3-8B (DPO)                        & 11/2024  & \texttt{allenai/Llama-3.1-Tulu-3-8B-DPO} & 1, 6 \\
Tulu-3-8B (RLVR)                & 11/2024  & \texttt{allenai/Llama-3.1-Tulu-3-8B} & 1, 6 \\
\bottomrule
\end{tabular}
\caption{Model suite used in this work.}
\label{tab:models}
\end{table}

\subsection{Embedding distance diagnostic}
\label{app:embedding-distance}
For $M=20{,}000$ sampled pairs of distinct token rows, let $d_m$ be the Euclidean distance between their L2-normalized residual activations and $\hat d_m$ the Euclidean distance between their saved UMAP embeddings. Pairs are drawn uniformly with seed 42; a row may occur in multiple pairs. We normalize the error by the mean original distance $\bar d=M^{-1}\sum_m d_m$:
\begin{align}
\mathrm{NRMSE}_{\mathrm{raw}} &= \frac{\sqrt{M^{-1}\sum_m(d_m-\hat d_m)^2}}{\bar d}
\end{align}

Fig.~\ref{fig:concept_quality_overview} reports this score.

\subsection{Concept annotation}
\label{app:annotation-prompts}
We use Mistral-Small-24B-Instruct-2501 to label each cluster from its up to 20 most frequent decoded tokens, four causal sentence contexts, and ten most frequent next-token predictions derived from the extraction corpus. Greedy decoding produces a 2--6-word English label. A separate prompt classifies the English concepts in Result 3 by type (semantic, syntactic, mixed, or unclear) and evidence orientation (input, output, both, or unclear). A third prompt classifies translation-shared concepts in Result 4. All three tasks use a single user message formatted with the labeler's chat template and greedy decoding (\texttt{do\_sample=False}). The generation limits are 30 tokens for concept names and 100 tokens for each classification task. Prompts, evidence, raw generations, and cleaned labels are retained. Labels are hypotheses for inspection that do not enter UMAP, HDBSCAN, CBA, or any other discovery metric. The R3 and R4 category summaries do depend on these annotations \citep{bills2023language}.

\subsubsection{Concept names}
This prompt supplies the descriptive labels used for concept inspection and matching examples. These names are not supplied to the R3 or R4 classifiers.
\begin{quote}
\small\raggedright
You are analysing the internal representations of a language model (\textless{}model identifier\textgreater{}, \textless{}number of layers\textgreater{} transformer layers total). You are looking at \textless{}layer description\textgreater{}. The tokens below come from \textless{}corpus description\textgreater{} text, split into subwords by its tokenizer (\textless{}tokenizer revision\textgreater{}).

The following subword tokens cluster together in the model's activation space at this layer --- meaning they trigger a similar internal representation pattern.

Input tokens: \textless{}top input tokens\textgreater{}

Predicted next tokens (what the full model predicts will follow these input tokens): \textless{}top predicted next tokens\textgreater{}

\textless{}depth-dependent note\textgreater{}

Example sentences:

- '\textless{}token\textgreater{}' in: \textless{}causal sentence prefix\textgreater{}

\textless{}up to four examples\textgreater{}

Task: Give a short label (2--6 words) naming what these tokens have in common.

Guidelines:

  \textbullet{} For morphological fragments (prefixes, suffixes, stems), describe their grammatical function rather than calling them 'fragments' or 'subwords' (e.g. 'verb suffix -ing', 'plural noun endings -s/-es', 'past tense -ed').

  \textbullet{} Avoid vague catch-alls such as 'abbreviations', 'subwords', 'legal terms', 'fragments', or 'German text' unless all tokens genuinely share that specific property.

  \textbullet{} Prefer a specific, distinctive label over a generic one.

  \textbullet{} Output ONLY the label --- no explanation, no punctuation at the end, no quotes.
\end{quote}
The layer description identifies the residual block and its relative depth. The next-token note varies with rounded relative depth, as follows; this is part of the labeling prompt, not a measured property of the layer.
\noindent\textit{Early (below 40\%).}
\begin{quote}
\small\raggedright
Note: at this early depth the predicted next tokens reflect the full model but may not align closely with the current representations --- use them as a secondary signal.
\end{quote}
\noindent\textit{Intermediate (40--69\%).}
\begin{quote}
\small\raggedright
Note: at this intermediate depth the predicted next tokens are moderately informative about the layer's function.
\end{quote}
\noindent\textit{Late (at least 70\%).}
\begin{quote}
\small\raggedright
Note: at this late depth the predicted next tokens are closely connected to the layer's representations --- weight them heavily when labelling.
\end{quote}
For boundary-only or empty decoded tokens, the prompt additionally states:
\begin{quote}
\small\raggedright
The visible token is only a standalone boundary/whitespace marker or otherwise decodes to an empty string. Use the causal sentence prefixes and predicted next tokens below to characterise the contextual representation captured by this cluster. Do not label it merely as an unknown concept.
\end{quote}
When begin-of-sequence markers dominate the displayed tokens, an additional guideline states:
\begin{quote}
\small\raggedright
Most tokens here are '\textless{}s\textgreater{}' (begin-of-sequence marker). Use the example sentences and predicted next tokens to characterise what type of content or article this cluster captures (e.g. 'Biographical article opening', 'Technical Wikipedia intro', 'Sentence-initial Die/The').
\end{quote}

\subsubsection{English concept type and evidence orientation (R3)}
The input contains up to 20 input tokens, ten predicted next tokens, and four contexts. Each context string is truncated to its first 200 characters by the classifier. Concept names, model identity, corpus description, and layer depth are withheld. Type and orientation are requested jointly with the following prompt.
\begin{quote}
\small\raggedright
You are classifying a concept cluster from a language model's internal representations using only the token and context evidence below.

Top tokens in cluster: \textless{}top input tokens\textgreater{}

Predicted next tokens: \textless{}top predicted next tokens\textgreater{}

Example sentences (cut at the token position):

  - '\textless{}token\textgreater{}' in: \textless{}causal sentence prefix\textgreater{}

\textless{}up to four examples\textgreater{}

Classify two independent properties.

concept\_type must be one of semantic, syntactic, mixed, unclear. Semantic means content-bearing meaning, topic, entities, or world knowledge. Syntactic means grammatical function, word form, punctuation, formatting, subword structure, or sentence position. Use mixed when both are material.

orientation must be one of input, output, both, unclear. Input means the current token and causal context best explain the cluster. Output means the predicted next-token evidence best explains it. Use both only when both evidence sources independently support the interpretation. The predicted tokens come from the final language-model head and are auxiliary evidence, not a direct readout from this intermediate layer.

Return only JSON in this form: \{"concept\_type":"\textless{}semantic|syntactic|mixed|unclear\textgreater{}","orientation":"\textless{}input|output|both|unclear\textgreater{}","reason":"brief evidence"\}
\end{quote}

\subsubsection{Bilingual shared-concept type (R4)}
This prompt is applied to concepts selected as translation-shared by the activation-frequency criterion in Result 4, independently of statistical significance. It uses up to 12 training tokens per language with counts and four representative parallel examples. Each sentence is truncated to 320 characters; these bilingual examples are sentence-level evidence, rather than the causal prefixes used above. Layer identity and shared-pair counts and rates are withheld. This task returns a type, a short English description, and a reason.
\begin{quote}
\small\raggedright
You are classifying a concept discovered jointly from two languages in a language model's residual-stream activations.

Languages: English and Mandarin Chinese

English training tokens: \textless{}token-count list\textgreater{}

Mandarin Chinese training tokens: \textless{}token-count list\textgreater{}

Representative parallel examples:

- English: \textless{}English sentence\textgreater{}

  Mandarin Chinese: \textless{}Mandarin sentence\textgreater{}

  activated tokens: English=[\textless{}tokens\textgreater{}], Mandarin Chinese=[\textless{}tokens\textgreater{}]

\textless{}up to four parallel examples\textgreater{}

Classify the organizing principle of this concept:

- semantic: shared meaning, topic, entity type, lexical meaning, or a translation-equivalent semantic category;

- syntactic: grammar, morphology, function words, punctuation, orthography, token fragments, or sentence position;

- mixed: both semantic and syntactic evidence are clearly central;

- unclear: the evidence is insufficient or incoherent.

Judge the activation evidence, not merely whether the two sentences are translations. Return exactly one JSON object with no markdown:

\{"classification":"semantic|syntactic|mixed|unclear","label":"2-6 word English description","reason":"one short English sentence"\}
\end{quote}

\subsection{Supplementary Results for Result~1: Sanity check}
Tab.~\ref{tab:nn-alignment-margin-per-model} shows the result of the sanity check stratified by model.

\begin{table}[ht]
\centering
\small
\setlength{\tabcolsep}{3pt}
\begin{tabular}{lcccccccc}
\toprule
 & \multicolumn{2}{c}{PCA (512-d)} & \multicolumn{2}{c}{CKA (linear)} & \multicolumn{2}{c}{CKA (RBF)} & \multicolumn{2}{c}{CBA (NLMCD)} \\
\cmidrule(lr){2-3} \cmidrule(lr){4-5} \cmidrule(lr){6-7} \cmidrule(lr){8-9}
Model & NN $\uparrow$ & $z$ $\uparrow$ & NN $\uparrow$ & $z$ $\uparrow$ & NN $\uparrow$ & $z$ $\uparrow$ & NN $\uparrow$ & $z$ $\uparrow$ \\
\midrule
GPT-2          & 1.000 & 1.407 & 0.917 & 0.461 & 1.000 & 1.288 & 1.000 & 1.893 \\
Llama3.1 8B Instruct & 0.938 & 1.827 & 1.000 & 1.702 & 1.000 & 1.694 & 1.000 & 2.546 \\
Llama3.1 8B Base     & 1.000 & 1.896 & 1.000 & 1.683 & 1.000 & 1.680 & 1.000 & 2.569 \\
Qwen3 4B       & 0.944 & 1.749 & 0.944 & 0.911 & 1.000 & 1.787 & 0.972 & 2.481 \\
Qwen3 8B       & 1.000 & 1.744 & 0.972 & 0.931 & 1.000 & 1.796 & 1.000 & 2.621 \\
Qwen3 32B      & 1.000 & 2.395 & 0.969 & 0.744 & 1.000 & 2.038 & 1.000 & 2.945 \\
Tulu SFT       & 1.000 & 1.898 & 1.000 & 1.715 & 1.000 & 1.714 & 1.000 & 2.551 \\
Tulu DPO       & 1.000 & 1.936 & 1.000 & 1.717 & 1.000 & 1.718 & 1.000 & 2.524 \\
Tulu final     & 0.969 & 1.953 & 1.000 & 1.716 & 1.000 & 1.716 & 1.000 & 2.504 \\
\bottomrule
\end{tabular}
\caption{Per-model nearest-neighbor alignment score and mean best-neighbor z-score ($z$) for all nine checkpoints underlying Tab.~\ref{tab:nn-alignment-margin}. NN measures whether the highest-scoring other layer is adjacent. $z$ measures how strongly the best immediate neighbor stands above the row mean relative to its standard deviation.}
\label{tab:nn-alignment-margin-per-model}
\end{table}

\subsection{Supplementary Results for Result~2: Layer-wise alignment}
\begin{figure}[htbp]
\centering
\setlength{\fboxsep}{0pt}
\setlength{\tabcolsep}{0pt}

\begin{subfigure}[b]{0.30\textwidth}
  \centering
  \includegraphics[width=\linewidth]{figures/cba_gpt2.png}
  \caption{GPT2}
  \label{fig:1a}
\end{subfigure}%
\hfill
\begin{subfigure}[b]{0.30\textwidth}
  \centering
  \includegraphics[width=\linewidth]{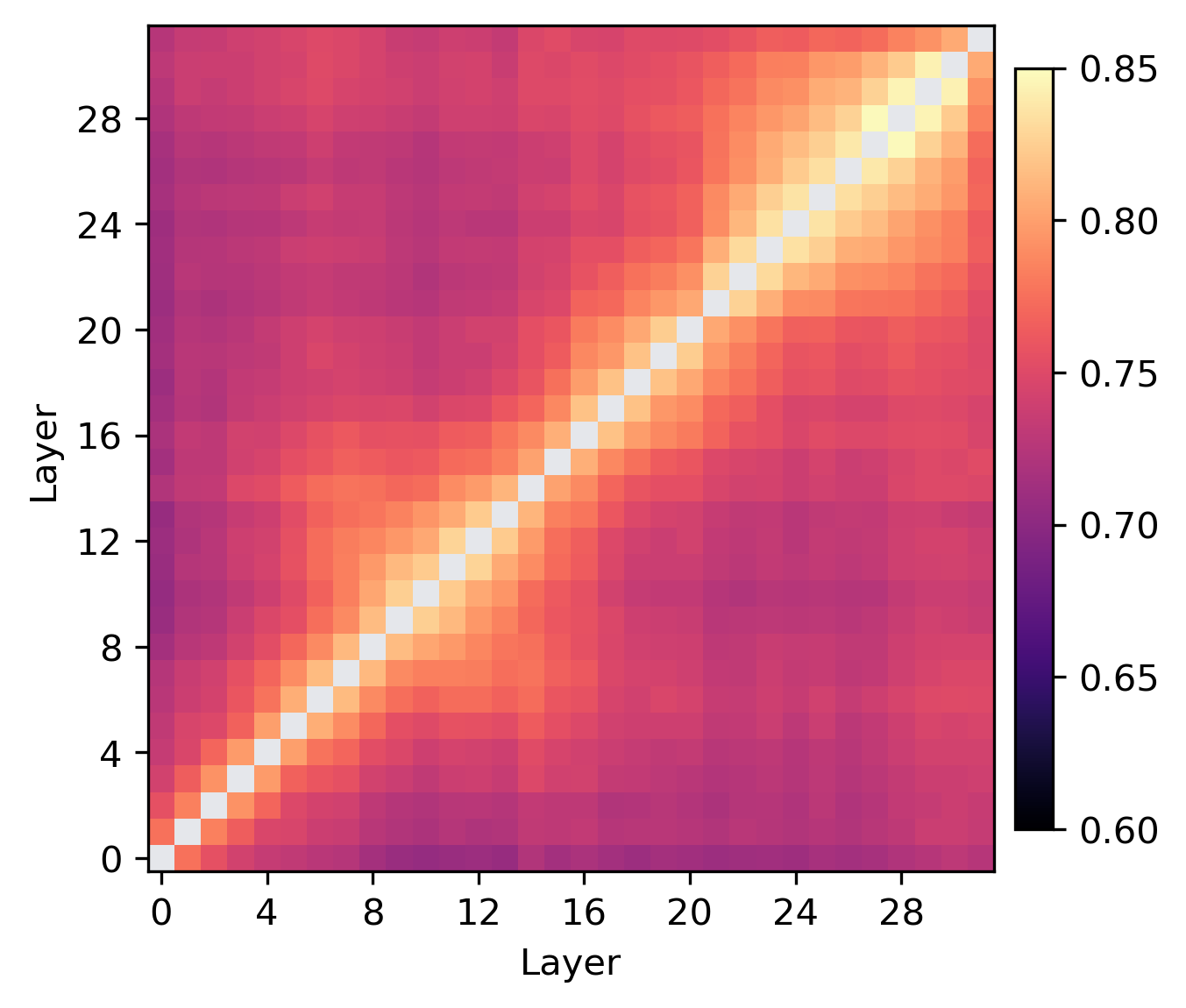}
  \caption{Llama3.1 8B Base}
  \label{fig:2}
\end{subfigure}%
\hfill
\begin{subfigure}[b]{0.30\textwidth}
  \centering
  \includegraphics[width=\linewidth]{figures/cba_llama8binstruct.png}
  \caption{Llama3.1 8B Instruct}
  \label{fig:3a}
\end{subfigure}

\vspace{0.5em}

\begin{subfigure}[b]{0.30\textwidth}
  \centering
  \includegraphics[width=\linewidth]{figures/cba_qwen4b.png}
  \caption{Qwen3 4B}
  \label{fig:4a}
\end{subfigure}
\hfill
\begin{subfigure}[b]{0.30\textwidth}
  \centering
  \includegraphics[width=\linewidth]{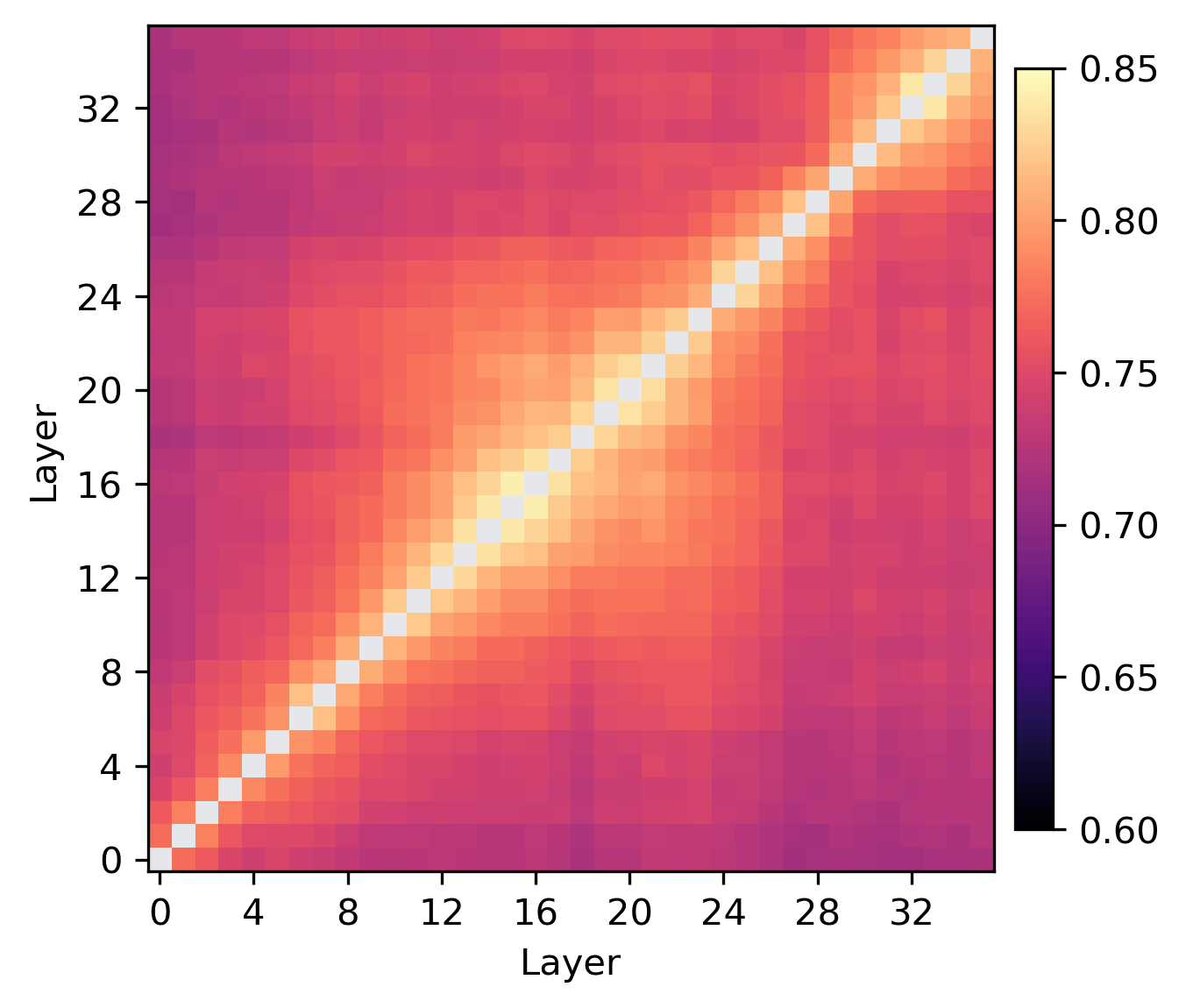}
  \caption{Qwen3 8B}
  \label{fig:5}
\end{subfigure}%
\hfill
\begin{subfigure}[b]{0.30\textwidth}
  \centering
  \includegraphics[width=\linewidth]{figures/cba_qwen32b.png}
  \caption{Qwen3 32B}
  \label{fig:6a}
\end{subfigure}%

\vspace{0.5em}

\caption{CBA layer alignment. Shared color scale $[0.60,0.85]$; self-alignment diagonals masked. Layer indices are zero-based. All panels use 5,000 samples.}
\label{fig:cba_grid_app}
\end{figure}

In Fig.~\ref{fig:cba_grid_app}, we show CBA layer-alignment matrices for six models under consideration. This supplements Fig.~\ref{fig:cba_grid} in the main text.

\subsection{Supplementary Results for Result~3: Descriptive/qualitative insights}
\begin{figure*}[!t]
  \centering
  \begin{subfigure}[t]{0.38\textwidth}
    \includegraphics[width=\linewidth]{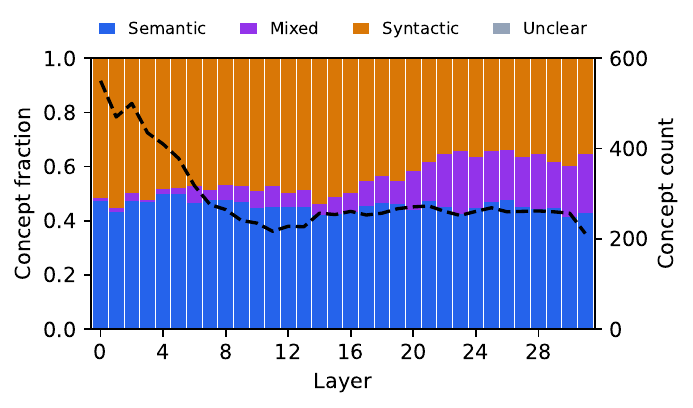}
    \caption{Llama-3.1-8B-Instruct: semantic/syntactic}
  \end{subfigure}\hspace{0.025\textwidth}
  \begin{subfigure}[t]{0.38\textwidth}
    \includegraphics[width=\linewidth]{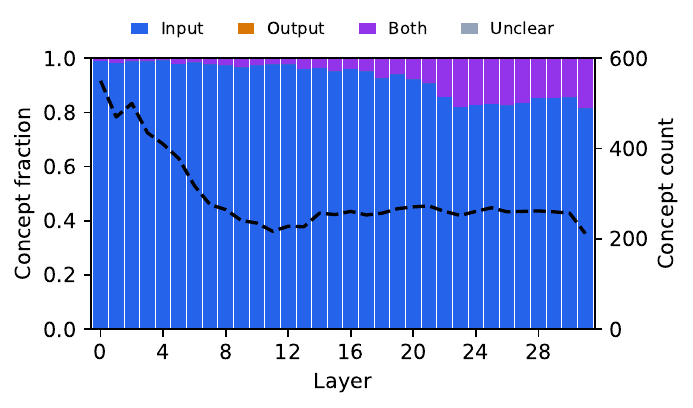}
    \caption{Llama-3.1-8B-Instruct: input/output}
  \end{subfigure}

  \caption{English concept categories and evidence orientation over residual depth. Bars show concept fractions (left axis). The dashed black line shows the total number of annotated concepts at each layer (right axis).}
  \label{fig:r3-app}
\end{figure*}

In Tab.~\ref{tab:r3-concept-annotations}, we show the summary statistics (summarized across all layers) of the descriptive concept analysis for Llama-3.1-8B-Instruct and Qwen3-8B. In Fig.~\ref{fig:r3-app}, we visualize the layer dependence of the concept properties for Llama-3.1-8B-Instruct, complementing the same analysis for Qwen3-8B in the main text. 

\begin{table*}[!t]
\centering
\small
\begin{tabular}{lrrrrrr}
\toprule
Model & Concepts & Semantic & Syntactic & Mixed & Input & Both \\
\midrule
Llama-3.1-8B-Instruct & 9,398  & 45.7\% & 45.1\% & 9.1\% & 93.4\% & 6.6\% \\
Qwen3-8B              & 11,496 & 39.8\% & 53.2\% & 7.0\% & 96.3\% & 3.7\% \\
\bottomrule
\end{tabular}
\caption{English concept annotations aggregated over residual layers. Percentages count concepts. Both denotes joint input/output evidence.}
\label{tab:r3-concept-annotations}
\end{table*}

\subsection{Supplementary Results for Result~4: Multilingual activation}
\begin{figure*}[!t]
  \centering
  \includegraphics[width=0.92\textwidth]{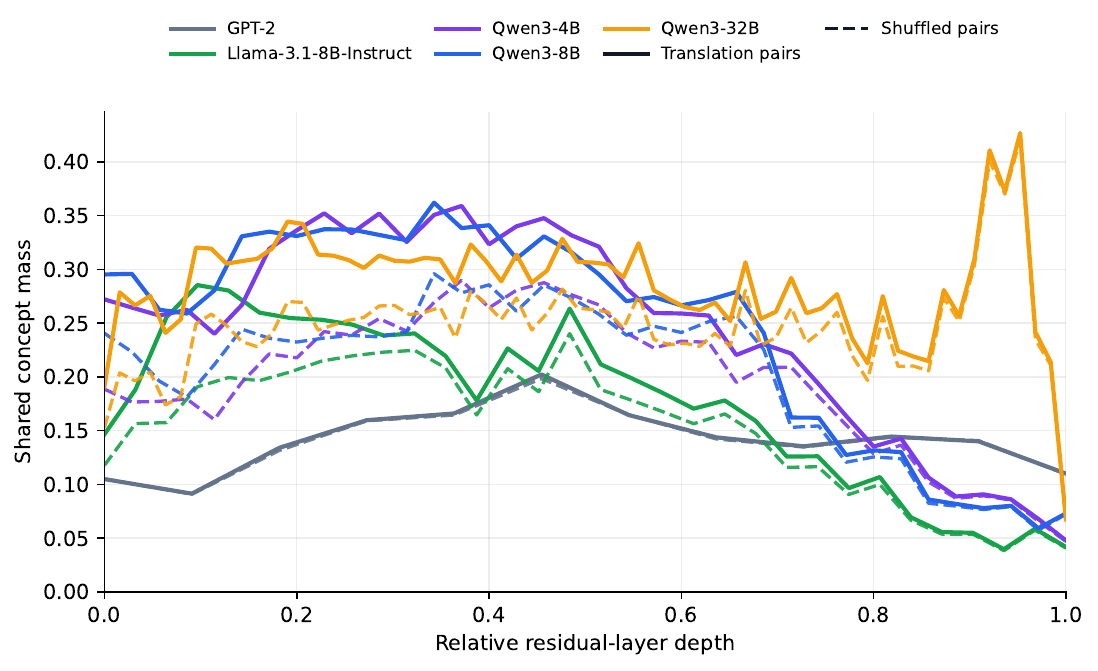}
  \caption{Shared concept mass over normalized residual depth for English--Mandarin translation pairs and shuffled-pair controls.}
  \label{fig:r4-parallel-activation}
\end{figure*}

Fig.~\ref{fig:r4-parallel-activation} summarizes shared concept mass as a function of the relative position of a layer in the architecture. In particular, it shows that in certain cases (such as Qwen3-32B close to the top layer), the token mass and the shuffled token mass largely coincide, indicating concepts that are always activated. This motivates the use of the difference between the original and the shuffled token mass as a measure for the multilingual activation.

\subsection{Supplementary Results for Result~6: Finetuning Case Study}
\begin{figure}[hbt]
  \centering
  \includegraphics[width=0.32\textwidth]{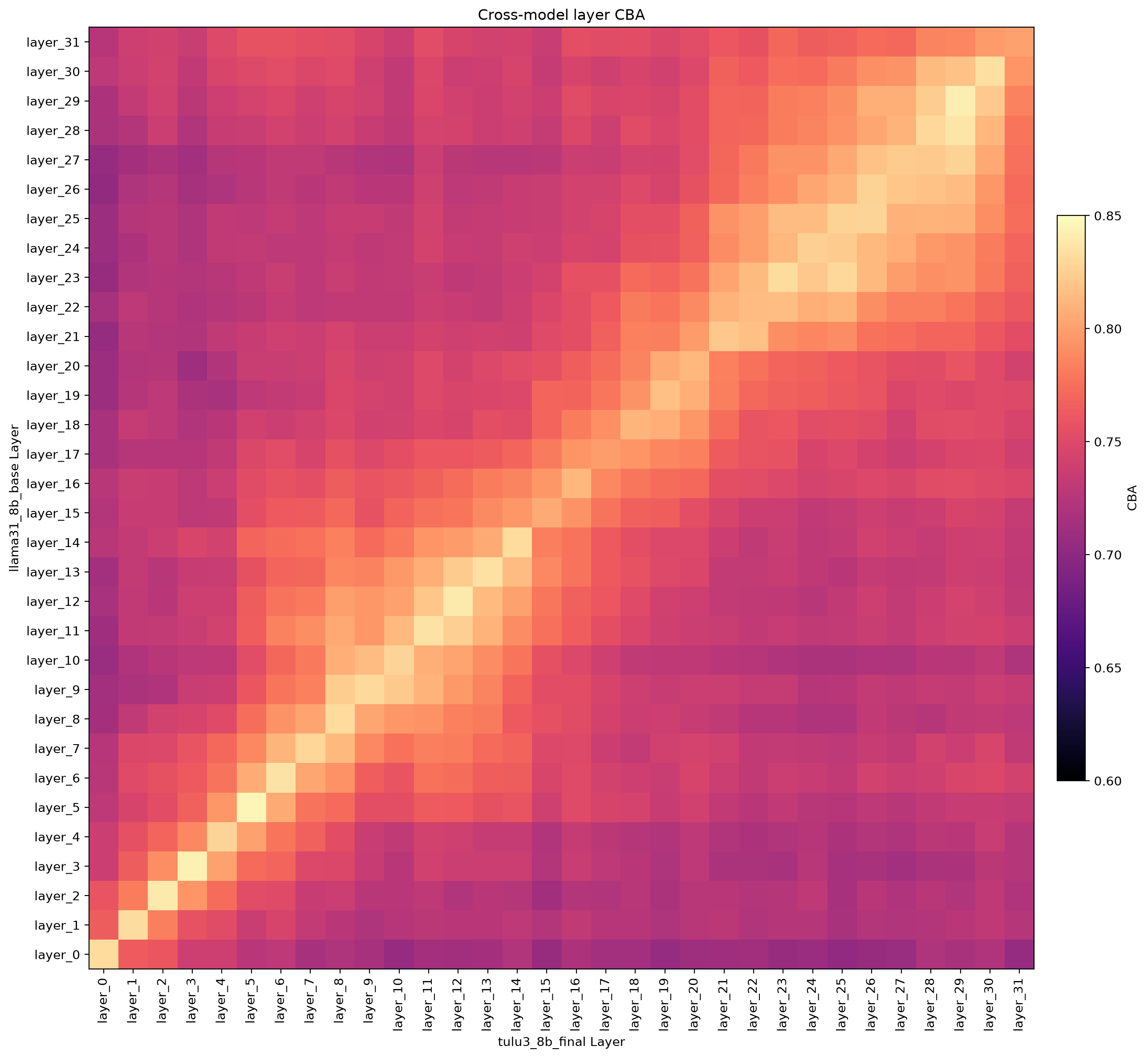}
  \hfill
  \includegraphics[width=0.32\textwidth]{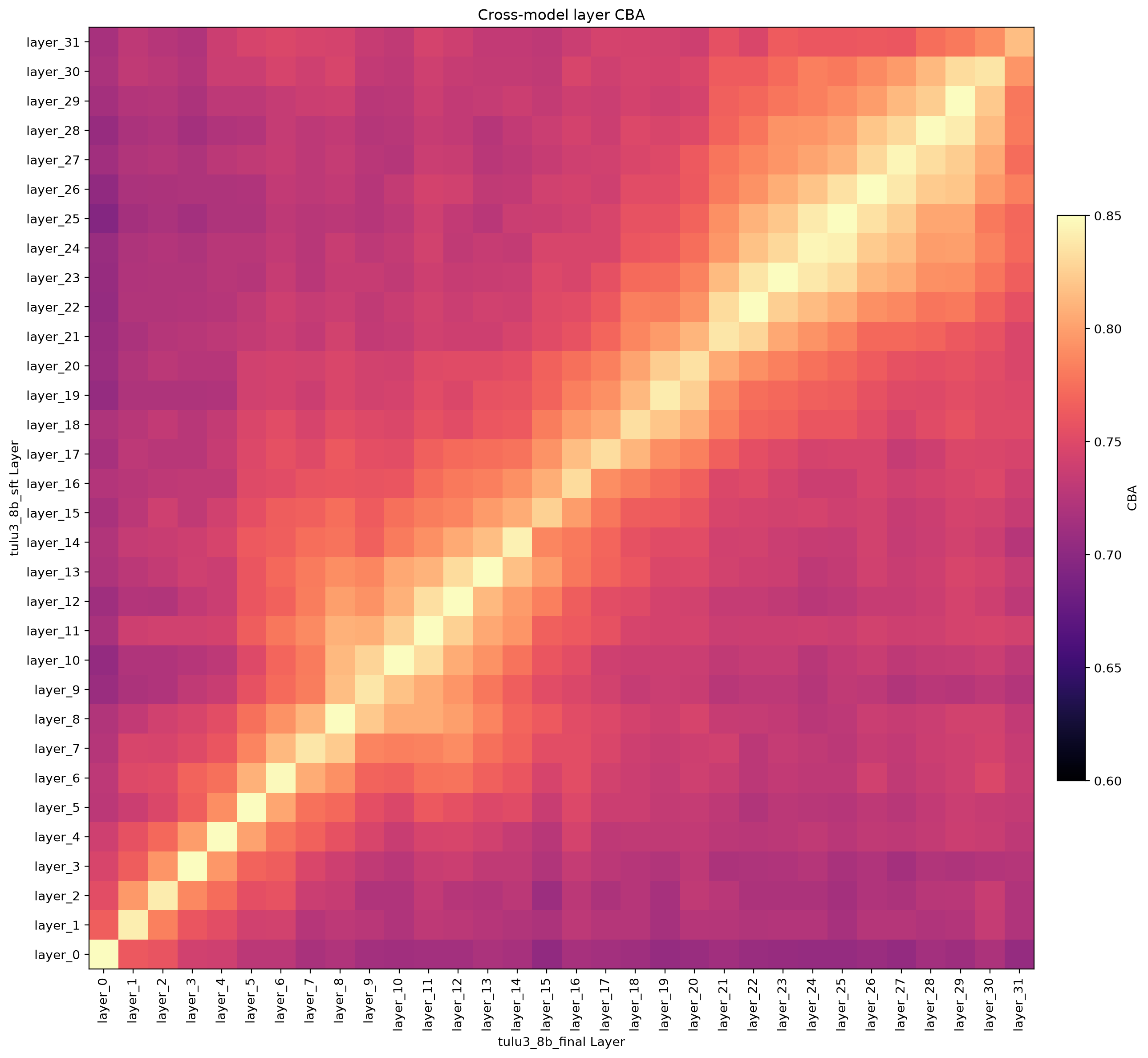}
  \hfill
  \includegraphics[width=0.32\textwidth]{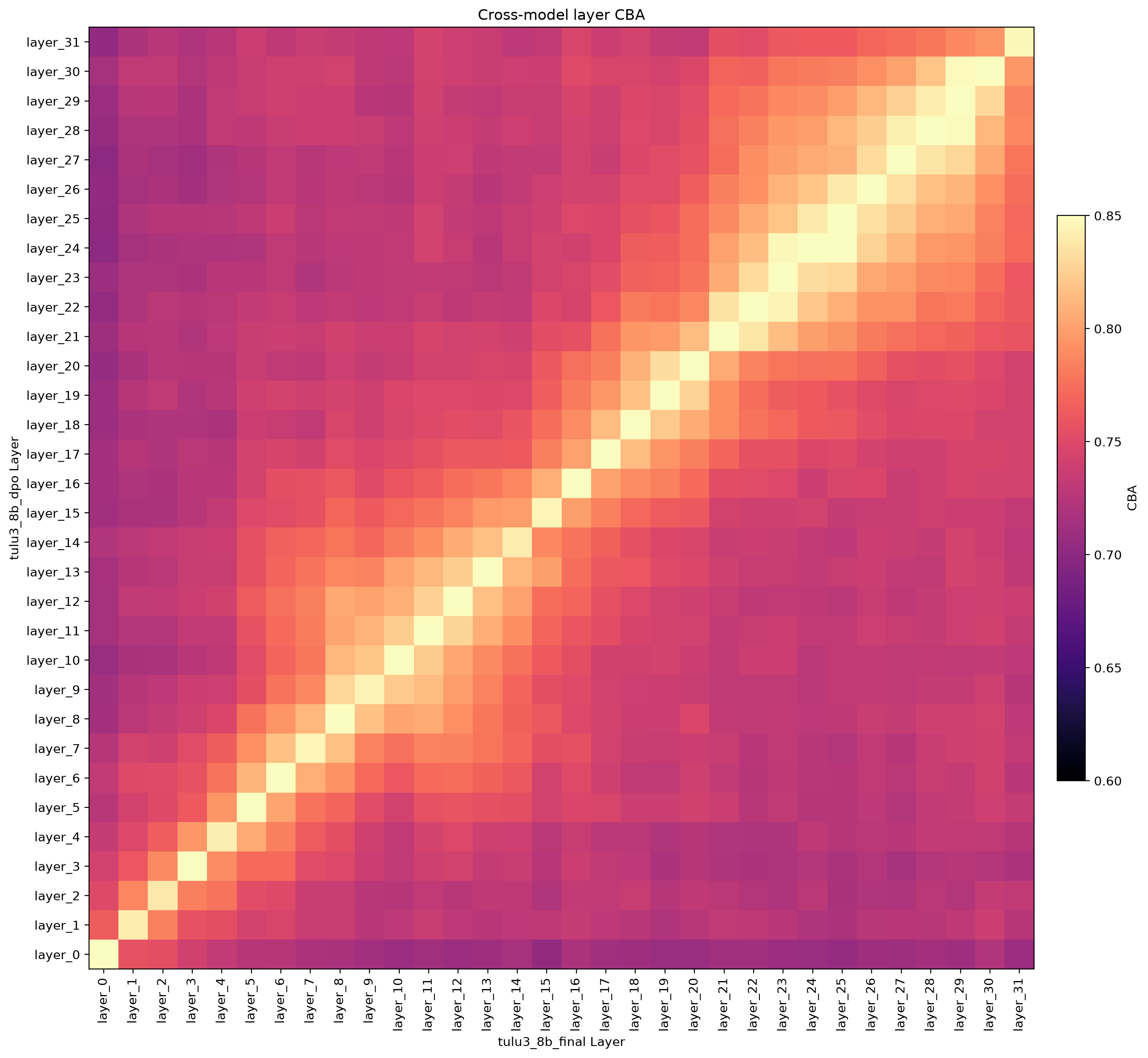}
  \caption{CBA Similarity Scores between finetune models. Left: Base vs. RLVR, Center: SFT vs. RLVR, Right: DPO vs. RLVR}
  \label{fig:CBA_models_finetune}
\end{figure}

\begin{figure}[hbt]
  \centering
  \includegraphics[width=0.48\textwidth]{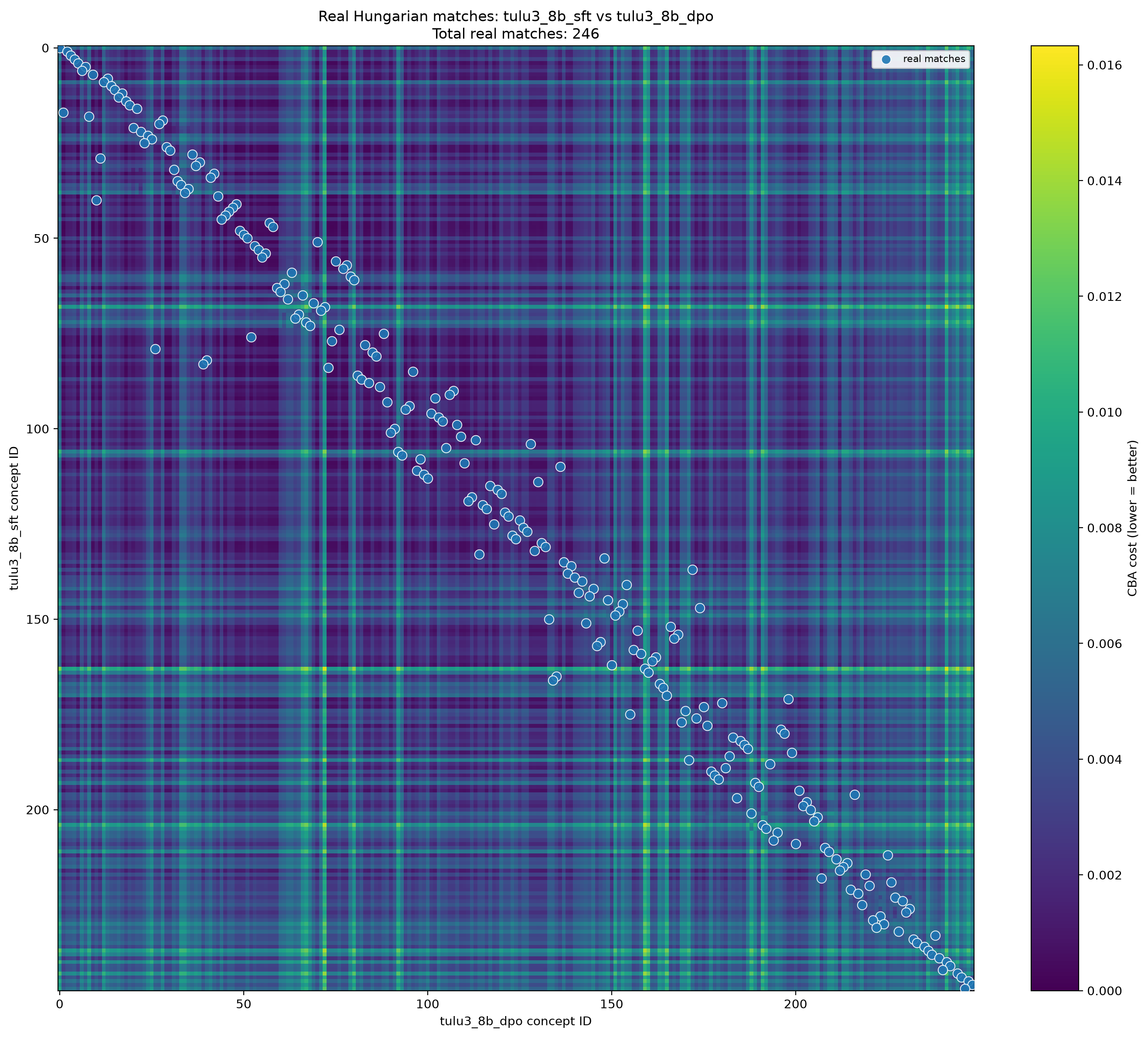}
  \hfill
  \includegraphics[width=0.48\textwidth]{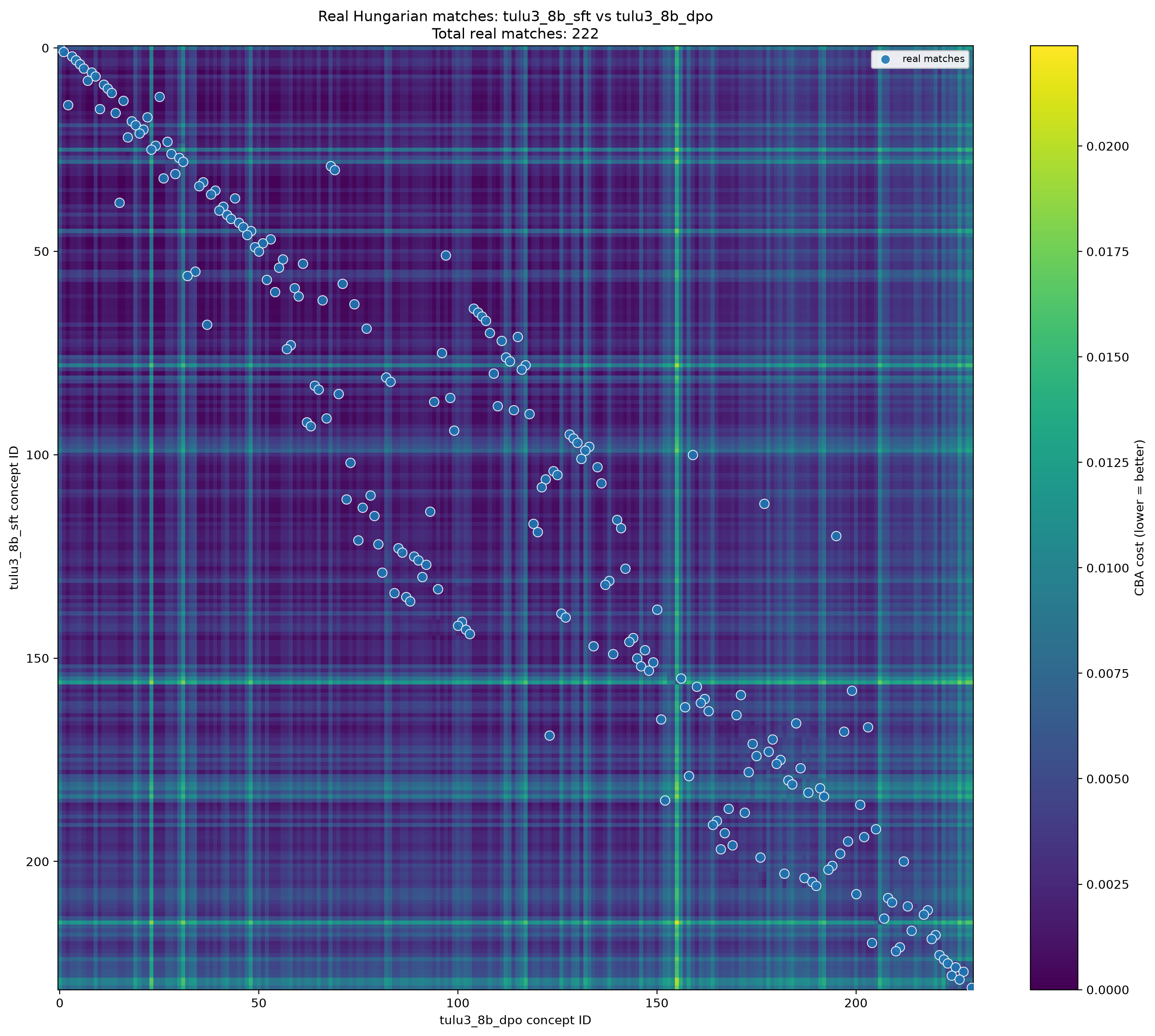}
  \hfill
  \caption{Pairwise-distance concept cost matrix and concept matches for SFT vs DPO. Left: Layer 14 Right: Layer 15}
  \label{fig:concept_cost_matrix_sft_dpo}
\end{figure}

In Fig.~\ref{fig:CBA_models_finetune}, we show CBA alignment matrices between different finetuned models. All of them show strong alignment of corresponding layers (diagonal). In Fig.~\ref{fig:concept_cost_matrix_sft_dpo}, we show a pairwise-distance concept cost matrix, which represents the basis for the application of the Hungarian algorithm for concept matching, as well as identified concept matches. In Tab.\ref{tab:match_examples_sft_dpo}, we show representative concept matches between SFT and DPO stages for two layers (14 and 15). Tab.\ref{tab:matches_combined} shows analogous matching results between the final Tulu model and the respective base model.

\begin{table*}[t]
\centering
\small
\setlength{\tabcolsep}{4pt}
\renewcommand{\arraystretch}{1.12}

\begin{threeparttable}
\caption{Representative Hungarian matches between SFT and DPO at Layers 14 and 15. Lower cost indicates greater similarity. ``Unmatched'' concepts are assigned to private dummy nodes. For all experiments the threshold for assigning matches to dummy nodes was set to 0.0025.}
\label{tab:match_examples_sft_dpo}

\begin{tabularx}{\textwidth}{@{}llXXr@{}}
\toprule
\textbf{Layer} & \textbf{Type} &
\textbf{SFT concept} & \textbf{DPO concept} &
\textbf{Cost} \\
\midrule

\multirow{10}{*}{14}
& Top        & section delimiters               & section delimiters               & $1.0\!\times\!10^{-4}$ \\
& Top        & Prepositions and conjunctions    & Prepositions and conjunctions    & $1.1\!\times\!10^{-4}$ \\
& Top        & Month names                      & Month names                      & $1.3\!\times\!10^{-4}$ \\
& Top        & sentence connectors              & sentence starters                & $1.4\!\times\!10^{-4}$ \\
& Top        & sentence starter                 & demonstrative pronouns           & $1.4\!\times\!10^{-4}$ \\
& Imperfect  & definite article                 & definite article                 & 0.0048 \\
& Imperfect  & prepositions and verbs           & Prepositional phrase start       & 0.0047 \\
& Imperfect  & physical structures              & places and structures            & 0.0045 \\
& Imperfect  & Last names                       & Surnames and Names               & 0.0044 \\
& Imperfect  & sentence initial the             & sentence initial capitalized the & 0.0043 \\
& Unmatched  & definite article, degree modifiers & Last names, military terminology, First names & --- \\
\midrule

\multirow{10}{*}{15}
& Top        & Prepositions and conjunctions    & Prepositions and conjunctions    & $0.6\!\times\!10^{-4}$ \\
& Top        & section delimiters               & section delimiters               & $0.7\!\times\!10^{-4}$ \\
& Top        & Name beginnings                  & Equality symbols                 & $0.7\!\times\!10^{-4}$ \\
& Top        & email symbols                    & email symbols                    & $0.8\!\times\!10^{-4}$ \\
& Top        & punctuation markers              & number separators                & $0.8\!\times\!10^{-4}$ \\
& Imperfect  & past tense verbs                 & past tense verbs                 & 0.0049 \\
& Imperfect  & Government Bodies                & Building materials               & 0.0045 \\
& Imperfect  & sentence start capitalized       & definite article                 & 0.0042 \\
& Imperfect  & common English prefixes          & consonant clusters               & 0.0042 \\
& Imperfect  & punctuation markers              & sentence boundary markers        & 0.0041 \\
& Unmatched  & Indefinite articles, Important Descriptors, conjunctions, auxiliary verbs, Prepositional start 
             & Common English words, subject pronouns, auxiliary verbs, auxiliary verbs and prepositions, Building and structure terms & --- \\
\bottomrule
\end{tabularx}

\end{threeparttable}
\end{table*}

\begin{table*}[t]
\centering
\small
\setlength{\tabcolsep}{4pt}
\renewcommand{\arraystretch}{1.12}

\begin{threeparttable}
\caption{Representative Hungarian matches between Llama-3.1-8B-Base and
Tulu-3-8B-Final at early, middle, and deep layers. Lower cost indicates
greater similarity. ``Unmatched'' concepts are assigned to dummy
nodes. For all experiments the threshold for assigning matches to dummy nodes was set to 0.0025.}
\label{tab:matches_combined}

\begin{tabularx}{\textwidth}{@{}llXXr@{}}
\toprule
\textbf{Layer} & \textbf{Type} &
\textbf{Llama concept} & \textbf{Tulu concept} &
\textbf{Cost} \\
\midrule

\multirow{8}{*}{0}
& Top        & Preposition \emph{in}              & Preposition \emph{in}              & $6.0\!\times\!10^{-5}$ \\
& Top        & Equality sign                      & Punctuation and digits              & $7.0\!\times\!10^{-5}$ \\
& Top        & Temporal conjunctions              & Temporal conjunctions               & $7.0\!\times\!10^{-5}$ \\
& Imperfect  & Punctuation comma                  & Sentence boundary                   & 0.0044 \\
& Imperfect  & Definite article                   & Sentence start                      & 0.0041 \\
& Imperfect  & Past-tense verb                    & Past-tense verb                     & 0.0040 \\
& Unmatched  & Article introducer; conjunction after; sentence boundary
                                                    & Verb \emph{to}; punctuation comma  & --- \\
\midrule

\multirow{8}{*}{16}
& Top        & Equality symbols                   & Equality symbols                    & $7.0\!\times\!10^{-5}$ \\
& Top        & Punctuation markers                & Month names                         & $1.3\!\times\!10^{-4}$ \\
& Top        & ISBN and numeric sequences         & Email addresses                     & $1.4\!\times\!10^{-4}$ \\
& Imperfect  & Sentence-ending punctuation        & Sentence-ending punctuation         & 0.0047 \\
& Imperfect  & Past and present verb forms        & Auxiliary verbs                     & 0.0045 \\
& Imperfect  & Past-tense action verbs            & Past-tense action verbs             & 0.0045 \\
& Unmatched  & Wikipedia article opening; sentence-boundary markers
                                                    & Punctuation markers; common English prefixes; surname beginnings
                                                                                         & --- \\
\midrule

\multirow{8}{*}{31}
& Top        & Lowercase prepositions             & Capitalization markers              & $9.0\!\times\!10^{-5}$ \\
& Top        & Capitalized names                  & Capitalized words                   & $1.8\!\times\!10^{-4}$ \\
& Top        & Cricket terminology                & Capitalized words                   & $1.9\!\times\!10^{-4}$ \\
& Imperfect  & Ordinal and temporal indicators    & Common English words                & 0.0048 \\
& Imperfect  & Common English words                & Common English words                & 0.0048 \\
& Imperfect  & Common English words                & Function words and metals           & 0.0046 \\
& Unmatched  & Year numbers; verb suffixes; verb forms and related words
                                                    & Year numbers; verb suffixes; initial consonants
                                                                                         & --- \\
\bottomrule
\end{tabularx}
\end{threeparttable}
\end{table*}

\end{document}